\documentclass[10pt,twocolumn,letterpaper]{article}

\usepackage[pagenumbers]{cvpr}            %
\usepackage[accsupp]{axessibility} %

\usepackage[dvipsnames]{xcolor}

\usepackage{epsfig}
\usepackage{graphicx}
\usepackage{comment}
\usepackage{amsmath,amssymb} %
\usepackage{xcolor}
\usepackage[export]{adjustbox}
\usepackage{bbm}

\usepackage[font=small,skip=5pt]{caption}
\usepackage{float}

\usepackage{multirow}
\usepackage{tabularx}
\usepackage{soul}
\usepackage{array}
\usepackage{comment}
\usepackage{outlines}

\usepackage[mathscr]{euscript}
\newcolumntype{L}[1]{>{\raggedright\let\newline\\arraybackslash\hspace{0pt}}m{#1}}
\newcolumntype{C}[1]{>{\centering\let\newline\\arraybackslash\hspace{0pt}}m{#1}}
\newcolumntype{R}[1]{>{\raggedleft\let\newline\\arraybackslash\hspace{0pt}}m{#1}}
\newcolumntype{Y}{>{\centering\arraybackslash}X}
\usepackage{booktabs}
\usepackage{enumitem}

\usepackage{amsmath}

\newcommand{\dataName}{ProciGen}
\newcommand{\modelName}{HDM}

\newcommand{\mat}[1]{\mathbf{#1}}

\newcommand{\vect}[1]{\mathbf{#1}}

\newcommand{\pose}[0]{\boldsymbol{\theta}}
\newcommand{\shape}[0]{\boldsymbol{\beta}}

\definecolor{ForestGreen}{RGB}{14,109,14}

\newcommand{\PCTwo}{$\text{PC}^2$}

\usepackage{multirow}

\definecolor{cvprblue}{rgb}{0.21,0.49,0.74}
\usepackage[pagebackref,breaklinks,colorlinks,citecolor=cvprblue]{hyperref}

\def\paperID{4173} %
\def\confName{CVPR}
\def\confYear{2024}

\title{Template Free Reconstruction of Human-object Interaction with Procedural Interaction Generation}

\author{
Xianghui Xie$^{1,2,3}$ 
\qquad
Bharat Lal Bhatnagar$^3$ 
\qquad
Jan Eric Lenssen$^3$ 
\qquad
Gerard Pons-Moll$^{1,2,3}$ \\
\\
{\small $^1$University of T\"ubingen, Germany \hspace{1cm} $^2$T\"ubingen AI Center, Germany  } \\
{\small $^3$Max Planck Institute for Informatics, Saarland Informatic Campus, Germany\hspace{1cm}} \\
{\small\href{https://virtualhumans.mpi-inf.mpg.de/procigen-hdm/}{https://virtualhumans.mpi-inf.mpg.de/procigen-hdm/}}\\
}

\begin{document}
\maketitle
\begin{abstract}
Reconstructing human-object interaction in 3D from a single RGB image is a challenging task and existing data driven methods do not generalize beyond the objects present in the carefully curated 3D interaction datasets.
Capturing large-scale real data to learn strong interaction and 3D shape priors is very expensive due to the combinatorial nature of human-object interactions. In this paper, we propose \dataName{} (Procedural interaction Generation), a method to procedurally generate datasets with both, plausible interaction and diverse object variation.
We generate 1M+ human-object interaction pairs in 3D and leverage this large-scale data to train our \modelName{} (Hierarchical Diffusion Model), a novel method to reconstruct interacting human and unseen object instances, without any templates. Our \modelName{} is an image-conditioned diffusion model that learns both realistic interaction and highly accurate human and object shapes.
Experiments show that our \modelName{} trained with \dataName{} significantly outperforms prior methods that require template meshes, and our dataset allows training methods with strong generalization ability to unseen object instances. Our code and data are released.

\end{abstract}
    
\section{Introduction}
\begin{figure}[t]
    \centering
    \includegraphics[width=\linewidth]{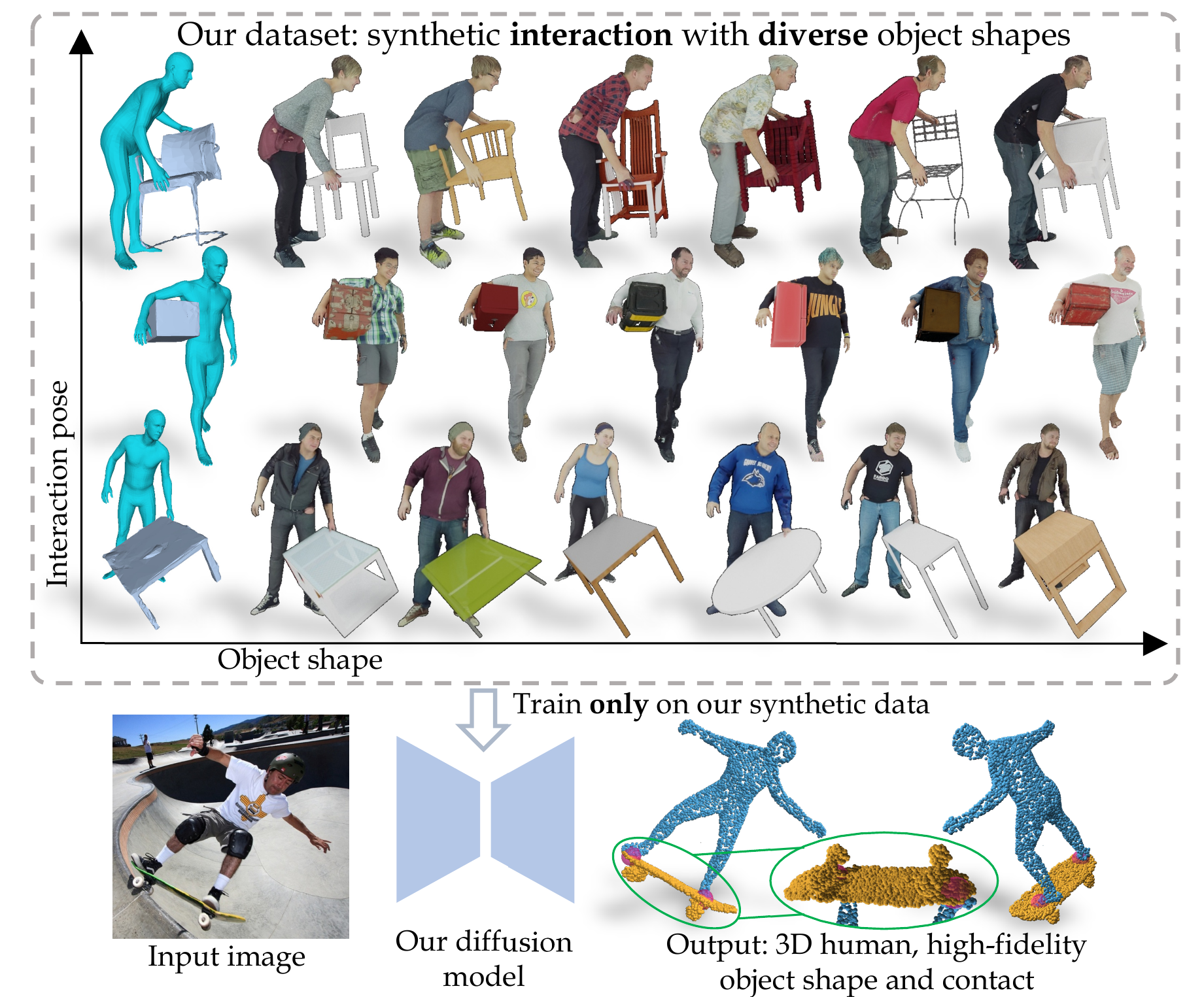}
    \caption{Given a single RGB image, our method trained \emph{only} on our proposed synthetic interaction dataset, can reconstruct the human, object and contacts, without any predefined template meshes. 
    }
    \label{fig:teaser}
\end{figure}
\label{sec:intro}
Modelling interactions between humans and their surroundings is important for applications like creating realistic avatars, robotic control and gaming. In this paper, we address the task of jointly reconstructing human and object from a monocular RGB image, without any prior object templates. This is very challenging due to depth-scale ambiguity, occlusions, diverse human pose and object shape variations. Data-driven methods have shown great progress in reconstructing humans \cite{rong2020frankmocap, pifuSHNMKL19, saito2020pifuhd, li2022cliff, hmrKanazawa17, he24nrdf, tiwari22posendf} or objects~\cite{liu2023zero1to3, genre} from monocular inputs thanks to large-scale datasets~\cite{Black_CVPR_2023bedlam, h36m_pami, Patel:CVPR:2021:AGORA, renderpeople, Zhang_2017_CVPR_BUFF, objaverse, shapenet2015, ModelNet_cvpr15}. However, methods for joint interaction reconstruction are still constrained by the amount of available data. Recent datasets like BEHAVE~\cite{bhatnagar22behave}, InterCap~\cite{huang2022intercap} capture real interactions with 10 to 20 different objects, which is far away from the number of objects in reality: the chair category from ShapeNet~\cite{shapenet2015} alone has more than 6k different shapes. Training on these real datasets has limited generalization ability to unseen objects (\cref{sec:exp-generalization}). Capturing real interaction data with more objects is prohibitively expensive due to the combinatorial nature: the number of humans times the number of objects leads to a huge number of variations. This motivates us to generate synthetic data which has been shown effective for pre-training reconstruction methods~\cite{Black_CVPR_2023bedlam, Patel:CVPR:2021:AGORA, hasson19_obman, pifuSHNMKL19, liu2023zero1to3}. 

Synthesizing realistic interaction for different objects is non-trivial due to variations of object topology, geometry details and complex interaction patterns. 
To address this, we propose \textbf{\dataName{}}: \underline{\bf Proc}edural \underline{\bf i}nteraction \underline{\bf Gen}eration, a method to generate interaction data with diverse object shapes. We design our method based on the key idea that the way humans interact with objects of the \emph{same} category is similar. And despite the geometry variations, one can still establish semantically meaningful correspondence between different objects. 
More specifically, we train an autoencoder to obtain correspondences between different objects from the same category, which are then used to transfer contacts from already captured human-object interactions to new object instances. Our method is scalable and allows the \emph{multiplicative combination} of datasets to generate over a million interactions with more than 21k different object instances, which is not possible via real data capture.

Current reconstruction methods~\cite{xie22chore,xie2023vistracker,bhatnagar22behave, melaskyriazi2023pc2} are not only bottle-necked by data. Template-based methods~\cite{xie22chore, xie2023vistracker, bhatnagar22behave} cannot generalize to unseen objects as they are \emph{trained only for specific object templates}. Template free methods like \PCTwo{}~\cite{melaskyriazi2023pc2}
cannot separate human and object, and have limited shape accuracy. See \cref{tab:comp-methods} for detailed comparison. To alleviate these issues, we propose \textbf{\modelName}: \underline{\bf H}ierarchical \underline{\bf D}iffusion \underline{\bf M}odel, that predicts accurate shapes and reasons about human-object semantics without using templates. Our key idea is to decompose the combinatorial interaction space into separate human and object sub-spaces while preserving the interaction context. We first use a diffusion model to jointly predict human, object and segmentation labels, and then use two separate diffusion models with cross attention that further refine the separate predictions.

We evaluate our data generation method \dataName{}, and model \modelName{}, on BEHAVE~\cite{bhatnagar22behave} and InterCap~\cite{huang2022intercap}. Experiments show that \modelName{} with \dataName{} significantly outperforms CHORE~\cite{xie22chore} (which requires object templates) and \PCTwo{}~\cite{melaskyriazi2023pc2}. Our \dataName{} dataset also significantly boosts the performance of \PCTwo{} and \modelName{}.
Methods trained on our synthetic \dataName{} dataset show strong generalization ability to real images even though the objects are unseen. 

In summary, our key contributions are:
\begin{itemize}
    \item We introduce the first procedural interaction generation method for synthesizing large-scale interaction data with diverse objects. With this, we generate 1M+ interaction images with 21k+ objects paired with 3D ground truth. 
    \item We propose a hierarchical diffusion model that can faithfully reconstruct human and object shapes from monocular RGB images without relying on template shapes. 
    \item Our dataset and code are publicly released. 
\end{itemize}

\begin{table}[t]
    \centering
    \scriptsize 
    \begin{tabular}{c c   c   c   c }
    \toprule[1.5pt]
         Method &  No-template & Shape acc. & General. & Semantic  \\
         \hline
         CHORE & \text{\sffamily X} & $\checkmark^*$& \text{\sffamily X} & \checkmark    \\
         \PCTwo{} & \checkmark & \text{\sffamily X} & \text{\sffamily X}  & \text{\sffamily X} \\
         \PCTwo{} + Our \dataName{} & \checkmark &  \text{\sffamily X} &  \checkmark & \text{\sffamily X} \\
         \hline
         Ours & \checkmark & \checkmark& \checkmark& \checkmark \\
    \bottomrule[1.5pt]
    \end{tabular}
    \caption{\textbf{Comparison of different reconstruction methods.} 
    CHORE \cite{xie22chore} reconstructs high shape fidelity with known template meshes but does not generalize to new object instances. \PCTwo{}~\cite{melaskyriazi2023pc2} is template-free but its shape predictions lack fidelity and generalization ability is constrained by existing datasets. Training \PCTwo{} with our \dataName{} dataset allows better generalization but it cannot reason contacts. 
    Our proposed data generation together with our hierarchical diffusion model can predict accurate shapes, generalize to unseen objects and reason about interaction semantics. 
    }
    \label{tab:comp-methods}
\end{table}

\section{Related Work}
\label{sec:related-work}

\textbf{Interaction Capture.} Modelling 3D interactions has been an emerging research field in recent years, with works that model hand-object interaction from RGB~\cite{GrapingField:3DV:2020,Corona_2020_CVPR,hasson19_obman, ehsani2020force, yang2021cpf}, RGBD~\cite{Brahmbhatt_2019_CVPR,Brahmbhatt_2020_ECCV, hampali2020honnotate} or 3D~\cite{zhou2022toch, zhou2024gears, GRAB:2020, GrapingField:3DV:2020, tendulkar2022flex} input, or predict contacts from RGB images~\cite{tripathi2023deco, chen2023hot, huang2022rich} and works that model human-scene interaction from single image~\cite{PROX:2019, Li_3DV2022MocapDeform, shimada2022hulc, Sandika_RAL23_HS_recon, yan2023cimi4d} or video \cite{Yi_MOVER_2022, CVPR21HPS}. A recent line of works model full body interacting with dynamic large objects~\cite{petrov2020objectpopup, zhang2022couch, guzov23ireplica, xu2021d3dhoi, liu2023hosnerf, bundlesdfwen2023, jiang2023instantnvr, jiang2023chairs, sun2021HOI-FVV, Han_2023_CHORUS, Kim_2023_NCHO}. BEHAVE~\cite{bhatnagar22behave} and follow up works~\cite{huang2022intercap, zhang2023neuraldome} capture interaction datasets, which allow training and benchmarking methods~\cite{xie2024rhobin} for reconstructing 3D human-object from single RGB images~\cite{zhang2020phosa, xie22chore, wang2022reconstruction} or videos~\cite{xie2023vistracker}. Despite impressive results, they require predefined mesh templates%
, which limits applicability to new objects. Our method is template-free and generalizes well to unseen objects. 

\textbf{Synthetic Datasets} are powerful resources to deep networks. For humans, synthetic rendering of 3D scans~\cite{renderpeople, bhatnagar2019mgn, tiwari20sizer, Patel:CVPR:2021:AGORA, yang2023synbody} are used extensively to train human reconstruction methods~\cite{pifuSHNMKL19, saito2020pifuhd, xiu2022icon, DoubleFusion2018, coronaLVD, bhatnagar2020ipnet, bhatnagar2020loopreg, xiu2023econ}. Recent work BEDLAM~\cite{Black_CVPR_2023bedlam} showed that training purely on synthetic datasets~\cite{Black_CVPR_2023bedlam, Patel:CVPR:2021:AGORA} allows strong generalization. Orthogonal to these, large scale 3D object CAD model datasets \cite{shapenet2015, ModelNet_cvpr15} are also used to pretrain backbone models\cite{jiang2023msp, yu2021pointbert, pang2022pointMAE, liu2022maskedPoints}. Other works~\cite{infinigen2023infinite, greff2021kubric, wrenninge2018synscapes, procthor} consider generating diverse scenes. While being useful for humans, objects or scenes respectively, they do not consider interactions. Our proposed approach can generate millions of interactions with diverse object shapes, allowing for training interaction reconstruction models with great generalization ability. 

\textbf{Diffusion-based Reconstruction.} Diffusion models \cite{ho2020ddpm, song19score_based} have been shown powerful for 3D reconstruction of human~\cite{huang2024tech, liao2023tada} and objects~\cite{liu2023zero1to3, seo2023_3dfuse, melaskyriazi2023pc2}. These works distil pretrained 2D diffusion model~\cite{melaskyriazi2023realfusion, zhou2023sparsefusion, poole2022dreamfusion, seo2023_3dfuse, ye2023featurenerf,huang2024tech, liao2023tada} or fine-tune diffusion model ~\cite{liu2023zero1to3, shi2023zero123plus, liu2023one2345} for 3D reconstruction from images. Recent works also propose image-conditioned point diffusion models for reconstruction~\cite{melaskyriazi2023pc2, Tyszkiewicz2023GECCOGP}. Despite remarkable results, they only model the distribution of single shapes, while our method can learn the complex interaction space with high shape fidelity.

\begin{figure*}[t]
    \centering
    \includegraphics[width=\linewidth]{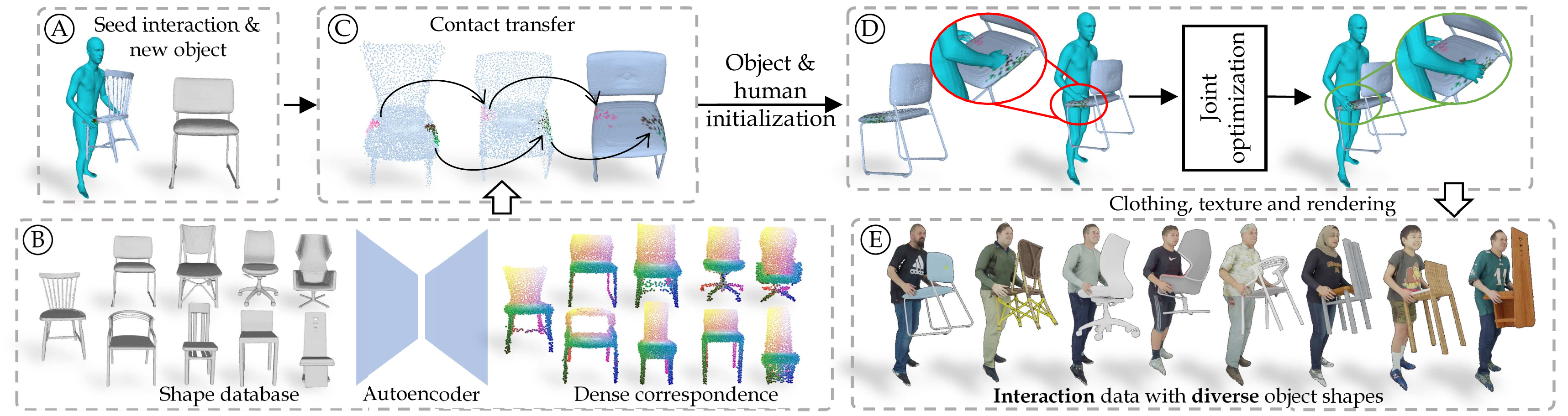}
    \caption{\textbf{Our procedural interaction generation method.} Given a seed interaction and a new object from the same category (A), we use a network to compute dense correspondences (B, \cref{subsec:dense_correspondence}), which allows us to transfer contacts and initialize the new object (C, \cref{subsec:contact-transfer}). We further optimize the human and object poses to avoid interpenetration while satisfying the transferred contacts (D, \cref{subsec:contact-opt}). We then add clothing and textures to render images, leading to a large interaction dataset with diverse object shapes (E, \cref{subsec:rendering}).  
    }
    \label{fig:data-synthesis}
\end{figure*}
\section{Method}
We first introduce our method to generate large amounts of interaction data with diverse object shapes in Sec.~\ref{sec:data-synthesis}. This data allows us to train our novel diffusion model with strong generalization ability, which is explained in \cref{sec:diffusion-model}.

\subsection{\dataName: Procedural Interaction Generation}\label{sec:data-synthesis}

Given a small seed dataset of captured human-object interactions and datasets of various object models, we aim to generate a large-scale interaction dataset with diverse object shapes. Via \emph{multiplicative scaling}, it would allow generating enormous data which is not possible by capturing real data. This is however non-trivial as object geometry varies strongly even within one category. Therefore, we propose a procedural method based on the key observation that humans interact similarly with objects of the same category. By transferring contacts from captured interactions to new object instances, we procedurally scale up the shape variations of real interaction datasets. The task involves solving four different sub-problems, as outlined in ~\cref{fig:data-synthesis}:

\begin{enumerate}
    \item \textbf{Establishing dense semantic correspondences} between all objects within one category (\cref{subsec:dense_correspondence}).
    \item \textbf{Transferring contacts} from real to synthetic objects, using the obtained correspondences  (\cref{subsec:contact-transfer}).
    \item \textbf{Jointly optimizing human and object} to the newly obtained contacts under a set of constraints (\cref{subsec:contact-opt}).
    \item \textbf{Rendering} novel intersection pairs with textures to make them available as training data (\cref{subsec:rendering}).
\end{enumerate}

 \subsubsection{Dense Semantic Correspondence}
 \label{subsec:dense_correspondence}
Given two meshes $\mathcal{M}$ and $\mathcal{M}'$ of two different objects of the same category, the problem of finding dense correspondence amounts to finding a bijective map $\psi: \mathcal{M} \rightarrow \mathcal{M}'$, which maps points from one mesh to their semantic counterparts on the other. In cases of arbitrary meshes with changing topology, this problem is heavily ill-posed~\cite{deng2021deformedIF, achlioptas2018learning_repre, Sundararaman2022DeformBasis}. Thus, we turn to an approximate solution on discrete surface samples that leverages the regularization and output ordering of MLPs\cite{liu2020learning_corr} and works well on a wide range of input topologies in practice.

Let $\{\mathcal{M}_i\}_{i=1}^M$ be a dataset of meshes from the same object category and $\mathbf{P}_i\in\mathbb{R}^{N\times 3}$ a point cloud sampled from the surface of $\mathcal{M}_i$. We train an autoencoder \mbox{$f: \mathbb{R}^{N\times 3}\mapsto \mathbb{R}^{N\times 3}$} on $\{\mathbf{P}_i\}_{i=1}^M$ to minimize the Chamfer distance between predicted and input point clouds. 
The network $f$ consists of a PointNet~\cite{Qi2016pointnet} encoder and a three-layer MLP decoder that takes unordered points as input and output ordered points. We found that the MLP decoder learns to reconstruct the objects as a mixture of low-rank point basis vectors, thus it automatically provides dense correspondence across objects through the order in the output, as also found in~\cite{zhou2022art,wewer2023simnp, Sundararaman2022DeformBasis}. 
Effective training of this network requires all shapes to be roughly aligned in a canonical space. When shapes are not aligned, we use ART~\cite{zhou2022art} which uses an additional network to predict an aligning rotation.  

To ensure the reconstruction quality, we overfit one network per object category. We show some example reconstructions and correspondences for chairs in \cref{fig:data-synthesis}B.

\subsubsection{Contact Transfer}\label{subsec:contact-transfer}
Given dense correspondences between a set of point clouds, we use them to transfer contact maps from one object to the other. Let $(\mathbf{H} \in \mathbb{R}^{M\times 3}, \mathbf{P} \in \mathbb{R}^{N\times 3})$ be a pair of human and object point clouds from an existing interaction dataset. And let $\mathbf{T} \in SE(3)$ be the non-rigid transformation that brings the object point cloud into canonical space where shapes are roughly aligned.
Then, we can find our \emph{contact set} as a set of point pairs from human and object that lie within a distance $\sigma$ to each other:
\begin{equation}
\mathcal{C}=\{(i,j) 
\mid ||\mathbf{H}_i-\mathbf{T}^{-1}f(\mathbf{T}\mathbf{P})_j)||^2_2<\sigma \} \textnormal{.}
\label{eq:contact-points}
\end{equation}
We first bring $\mathbf{P}$ into canonical pose, then apply $f$ to obtain a coherent point cloud, which is brought back into interaction pose by $\mathbf{T}^{-1}$. Since our autoencoder $f$ produces coherent point clouds, the obtained contact set can be directly transferred to all other objects $\mat{P}^\prime$ within the category, allowing us to pair the human with all other objects, one example transfer is shown in \cref{fig:data-synthesis}C. Once we transfer the contact points to the new object, we can find the corresponding contact facets in the meshes that have the smallest distances.

\subsubsection{Contact-based Joint Optimization}\label{subsec:contact-opt}
The newly obtained contact sets define how and where a human should interact with the new object. We can also transform the object from canonical to interaction pose with our dense correspondence. 
However, this naive placement does not guarantee the plausibility of the interaction due to object geometry changes (see~\cref{fig:data-synthesis}D). Hence, we propose a joint optimization to refine the human and object pose such that: \textbf{a)} contact points are close to each other, \textbf{b)} contact face normals match, and \textbf{c)} interpenetration is avoided. 

We use the SMPL-H~\cite{smpl2015loper,MANO:SIGGRAPHASIA:2017} body model $H(\pose, \shape)$ to parameterize the human as a function of pose $\pose$ and shape $\shape$ parameters. The object pose is given as non-rigid transformation $\mathbf{T}\in SE(3)$, and we denote the new object point cloud to which we have transferred contacts as $\mathbf{P}^\prime$. We find the refined human-object poses jointly, by minimizing:
\begin{equation}
    L(\pose, \shape, \mathbf{T}) = \lambda_c L_c  + \lambda_n L_n + \lambda_\text{colli} + \lambda_\text{init} L_\text{init} \textnormal{,}
    \label{loss-joint-opt}
\end{equation}
where the individual loss terms are given as:
\begin{itemize}
    \item \textbf{Contact:} $L_c=\sum_{(i, j)\in \mathcal{C}}||\mathbf{H}_i - \mathbf{P}_j^\prime||^2_2$, minimizing the distance between contact points.
    \item \textbf{Normal:} $L_n=\sum_{(i, j)\in \mathcal{C}}||1+\vect{n}_i^T \vect{n}_j||^2_2$, ensuring that normals $\vect{n}_i, \vect{n}_j$ of contacting faces point in opposite directions.
    \item \textbf{Interpenetration:} $L_\text{colli}$ penalizing interpenetration based on the bounding volume hierarchy \cite{Tzionas2016capture-handobject}.
    \item \textbf{Initialization:} $L_\text{init}$ is the L2 distance between new and original human pose, regularizing the deformation.
\end{itemize}
The pose $\pose$ is initialized from the original human pose and $\shape$ is randomly sampled from a set of registered scans~\cite{bhatnagar2019mgn}. The object pose $\mat{T}$ is initialized by Procrustes alignment between the two coherent point clouds $\mat{P}^\prime$ and $\mat{P}$. 
After joint optimization we obtain realistic interactions, see \cref{fig:data-synthesis}D.

\subsubsection{Dataset Rendering}\label{subsec:rendering}
Our contact transfer and joint optimization provide us the skeleton of interaction with new objects. To render them as images, we take the optimized SMPL-H parameters from \cref{subsec:contact-opt} and randomly sample the clothing deformation and texture from SMPL+D registrations in MGN~\cite{bhatnagar2019mgn}. For objects, we use the original texture paired with the mesh. We render the scenes in Blender~\cite{blender}, which is detailed in supplementary. See example renderings in \cref{fig:data-synthesis}E.

\textbf{Method Scalability.} %
We emphasize that the proposed procedural generation is a scalable solution that can generate large-scale datasets with only a small amount of effort for data capture: with 2k different interactions (e.g. BEHAVE~\cite{bhatnagar22behave} chair interaction), 6k different objects (e.g. Shapenet chairs~\cite{shapenet2015}) and 100 human scans (e.g. MGN~\cite{bhatnagar2019mgn}), one can have maximum \emph{1.2 billion} different variations in total, which is not possible with real data capture.
The data scale allows for training powerful models that reach performance not obtainable by training on real data only. An example of such a method is detailed in the next section.

\subsection{HDM: Hierarchical Diffusion Model}\label{sec:diffusion-model}
\begin{figure*}[t]
    \centering
    \includegraphics[width=\linewidth]{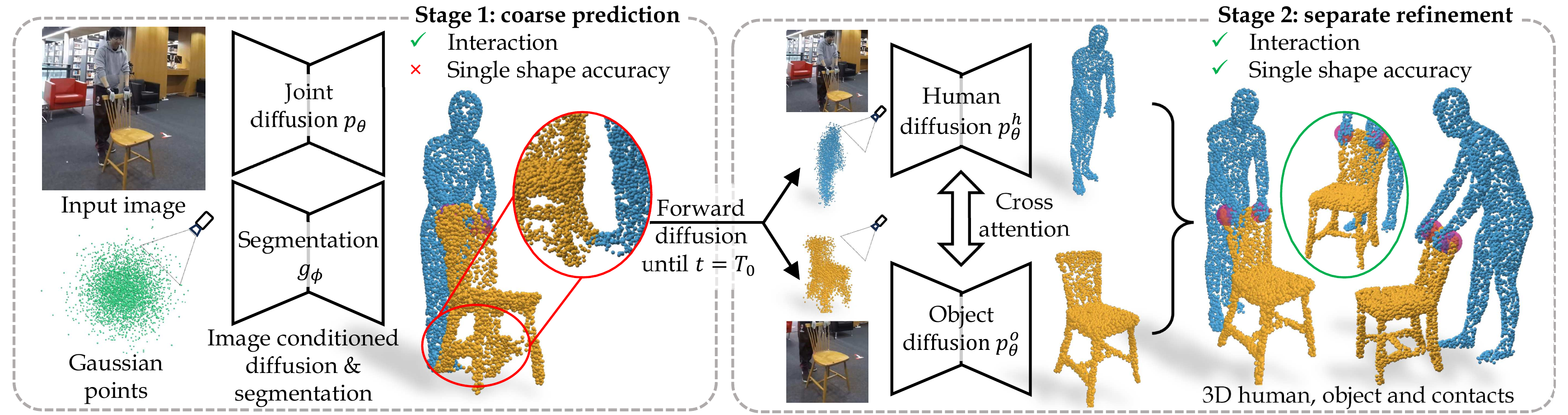}
    \caption{\textbf{Our hierarchical diffusion model.} Given an RGB image of a human interacting with an object, we first jointly reconstruct the human and object as one point cloud with segmentation labels (Stage 1, \cref{subsec-joint-diffusion}). This prediction reasons interaction but lacks accurate shapes. We then use two diffusion models for human or object separately with cross attention to refine the initial noisy prediction while preserving the interaction context(Stage 2, \cref{subsec:hierarchical-diffusion}). Our hierarchical design faithfully predicts interaction and shapes. 
    }
    \label{fig:method-hierarchical-diffusion}
\end{figure*}

Modelling the joint shape space of humans interacting with objects is difficult since the product of human and object shape variations is huge. One solution is to use two separate networks that reconstruct human and object respectively. 
However, such a method ignores the interaction cues that have been show important for coherent reconstruction~\cite{bhatnagar22behave, xie22chore, xie2023vistracker, Yi_MOVER_2022}. This motivates us to design a hierarchical solution where we first jointly estimate both human and object(\cref{subsec-joint-diffusion}), and then use separate networks that focus on refining individual shape details (\cref{subsec:hierarchical-diffusion}). An overview of our method can be found in \Cref{fig:method-hierarchical-diffusion}. 

\subsubsection{Preliminaries}

\textbf{Task Overview.} Given an input RGB image $\mat{I}$ of a person interacting with an object, we aim to jointly reconstruct 3D human and object point clouds $\mat{P}^h, \mat{P}^o$. Same as prior works~\cite{xie22chore, zhang2020phosa, xie2023vistracker}, we assume known 2D human and object masks, which we consider a weak assumption, given recent advances in 2D segmentation~\cite{sam_hq, kirillov2023sam, fbrs2020}. 

Due to the ambiguity from monocular input, we adopt a probabilistic approach for 3D reconstruction, which has been proven effective in learning multiple modes given same input \cite{melaskyriazi2023pc2, zhang2022relpose, zhou2023sparsefusion}. Specifically, we use a diffusion model \cite{ho2020ddpm} to learn the distribution of 3D human object interactions conditioned on a single image.

\textbf{Diffusion models}~\cite{ho2020ddpm, song19score_based} are general-purpose generative models that consist of iterative forward and reverse processes. Formally, given a data point $\vect{x}_0$ sampled from a data distribution $p_\text{data}$, the forward process iteratively adds Gaussian noise $q(\vect{x}_t| \vect{x}_{t-1})$ to the sample $\vect{x}_0$. 
The distribution at step $t$ can be computed as: 
\begin{align}
    \vect{x}_t &= \sqrt{\Bar{\alpha}_t}\vect{x}_0+\epsilon\sqrt{1-\Bar{\alpha}_t}
    \label{eq:diffusion-forward}
\end{align}
where $\Bar{\alpha}_t$ controls the noise level at step $t$ and $\epsilon\sim\mathcal{N}(\mat{0}, \mat{1})$~\cite{ho2020ddpm}.  
The reverse process starts from Gaussian noise at step $T$ and gradually denoises it back to the original data distribution $p_\text{data}$ at step 0. At each reverse step, we use a neural network $p_\theta$ to approximate the distribution: $p_\theta\approx q(\vect{x}_{t-1}|\vect{x}_t)$. The network is trained with the variational lower bound to maximize the log-likelihood of all data points, which is parametrized to minimize the L2 distance between the true noise $\epsilon$ and network prediction\cite{ho2020ddpm}:
\begin{equation}
    \mathcal{L} = E_{t\sim [1, T]}E_{\epsilon_t\sim \mathcal{N}(0, \mat{I})}[||\epsilon_t - p_\theta(\vect{x}_t, t)||^2_2]
    \label{eq:diffusion-loss}
\end{equation}

\subsubsection{Joint Human-object Diffusion}\label{subsec-joint-diffusion}
In this first stage, we simultaneously predict both human and object hence the output is one point cloud $\mat{P}\in \mathbb{R}^{N\times 3}$. We adopt \PCTwo{}~\cite{melaskyriazi2023pc2} that diffuses point cloud conditioned on single images. Formally, we use a point voxel CNN~\cite{Zhou_2021_ICCV_PVD, liu2019pvcnn} $p_\theta: \mathbb{R}^{N\times D}\mapsto \mathbb{R}^{N\times 3}$ as the point diffusion model. Here $D$ is the feature dimension. To obtain per-point input features, we first use a pre-trained encoder~\cite{he_mae_2022} to extract feature grid $\mat{F}\in \mathbb{R}^{F\times H^\prime \times W^\prime}$ from input image $\mat{I}$, here $F$ and $H^\prime, W^\prime$ are feature and spatial dimensions respectively. Points $\vect{p}\in \mat{P}$ are then projected to 2D image plane with $\pi(\cdot):\mathbb{R}^3\mapsto \mathbb{R}^2$ to extract pixel-aligned feature $\mat{F}_{\pi(\vect{p})}$. We further concatenate it with point location and diffusion timestamp encodings $t_\text{enc}$ as the input to the diffusion model: $\mat{F}_{\vect{p}}=(\mat{F}_{\pi(\vect{p})}, \vect{p}, t_\text{enc})$. To allow generative prediction for points that are occluded, the image features $\mat{F}_{\pi(\vect{p})}$ are set to zeros when points are occluded~\cite{melaskyriazi2023pc2}.

\subsubsection{Hierarchical Diffusion for Interaction}\label{subsec:hierarchical-diffusion}
Naively using one network to reconstruct interaction leads to noisy point predictions (see \cref{fig:generalization-comp}), as the combinatorial shape space of human-object interaction is too complex to model. 
Thus, we propose a second stage to refine human and object shapes separately, by having two additional diffusion models while also preserving the interaction context. In the following, we discuss special aspects of our second stage, namely \textbf{1)} how the point cloud is segmented into human and object, \textbf{2)} how separate networks are designed to model interaction, \textbf{3)} how these models are combined.

\textbf{Point cloud segmentation.} \label{para-segmentation}
To reason the contacts during interaction and obtain accurate shapes for human and object separately, the combined point cloud needs to be segmented into the points for human and object. 
To this end, we use an additional network $g_\phi: \mathbb{R}^{N\times D}\mapsto \{0, 1\}^{N}$ that takes point features $\mat{F}_\vect{p}$ as input and predicts a binary label to indicate whether this is a human or object point. With this prediction, we can segment the point cloud $\mat{P}$ predicted by $p_\theta$ into human and object points $\mat{P}^h, \mat{P}^o$. 

\textbf{Preserving interaction context.}\label{para-cross-attetion} In our second stage, we use two additional diffusion models $p_\theta^h, p_\theta^o$ to predict human and object. The networks follow the same design as the joint network $p_\theta$ using PVCNN \cite{Zhou_2021_ICCV_PVD, liu2019pvcnn}.
To encourage the networks to explore interaction cues, we add cross-attention layers between the encoder and decoder layers of human and object branches. Given downsampled points $\mat{P}_{l}\in \mathbb{R}^{N_{l}\times 3}$ with features $\mat{F}_{l}\in \mathbb{R}^{N_{l}\times D_{l}}$ after network layer $l$, we propagate information from human branch to object branch by computing feature: 
\begin{equation}
    \mat{F}_l^{h\mapsto o} = \text{Attn}(\text{enc}(\mat{P}_l^o), \text{enc}(\mat{P}_l^h), \mat{F}_{\mat{P}^h_l}), 
    \label{eq:cross-attention}
\end{equation}
where $\text{Attn}(\mathbf{Q}, \mathbf{K}, \mathbf{V})$ is learnable cross attention\cite{NIPS2017_attention}, $\text{enc}(\cdot)$ is positional encoding from NeRF~\cite{mildenhall2020nerf}, and $\mat{F}_{\mat{P}^h_l}=(\text{enc}(\mat{P}_l^h), \mat{C})$ is the concatenation of positional encoding and onehot encoding $\mat{C}$ indicating these points belong to human. The attention feature $\mat{F}_l^{h\mapsto o}$ is then concatenated to the object feature $\mat{F}_{l}^o$ as input to the next layer. We propagate information from object to human branch similarly. 
\begin{figure*}[t]
    \centering
    \includegraphics[width=0.98\linewidth]{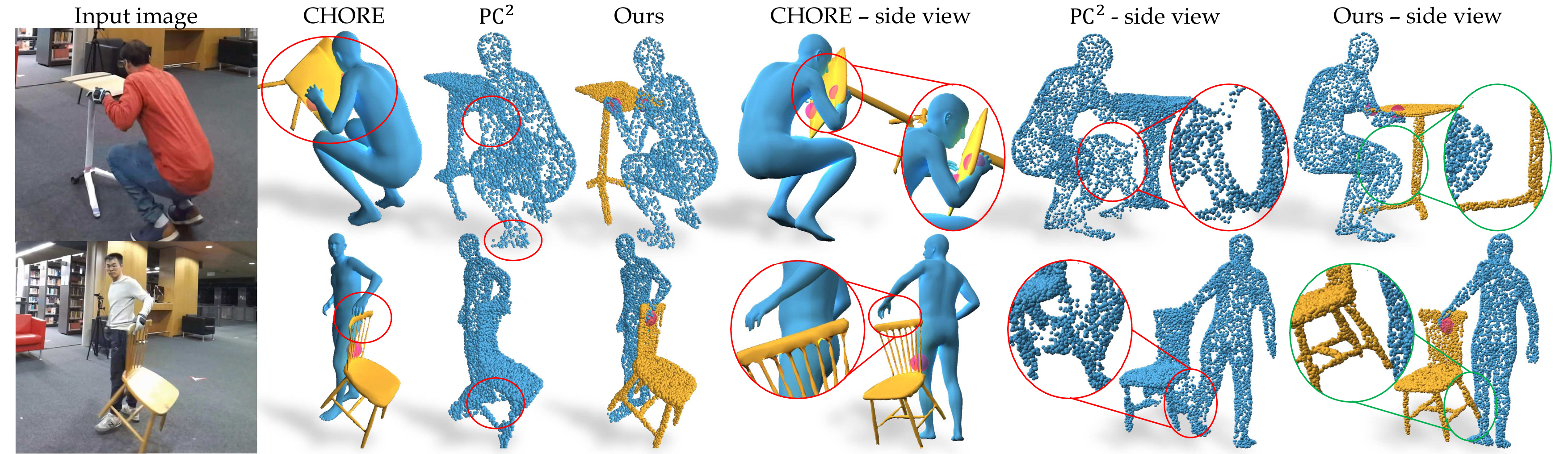}
    \caption{\textbf{Comparing reconstruction results on BEHAVE\cite{bhatnagar22behave} dataset.} CHORE\cite{xie22chore} relies on object mesh templates and the prediction is inaccurate for challenging poses. \PCTwo{}\cite{melaskyriazi2023pc2} does not rely on templates but its predicted point clouds are noisy (red circles) and it cannot predict contacts. Ours can reason about human object interaction, and predicts high-fidelity human and object shapes without templates. }
    \label{fig:results-comparison}
\end{figure*}

\indent\textbf{Model Combination.}\label{para-model-combination} With the separate networks $p_\theta^h, p_\theta^o$, one can run the full reverse diffusion process from $t=T$ to $t=0$ and then combine the denoised points. However, this does not leverage the predicted interaction context from the joint reconstruction stage and is slow. We hence start the reverse diffusion steps from an intermediate step $t=T_0$ instead of $T$. Specifically, after denoising and segmentation with the joint model, we apply the forward diffusion process to $\mat{P}^h$ and $\mat{P}^o$ until step $t=T_0$ using \cref{eq:diffusion-forward}. Then, the individual diffusion models take the noised points as input and gradually denoise them until step $t=0$. 
The forward process destroys local noisy predictions but keeps the global structure of human-object interaction. We set $T_0=\frac{T}{2}$, see supp. for analysis of this value. Our hierarchical design is important to obtain sharp predictions, see \cref{table:ablate-hdm} and \cref{fig:generalization-comp}. 

Recall from \cref{eq:diffusion-forward} that the forward diffusion ends up with a normal distribution. Hence the input and output points of all diffusion models are centered at the origin and scaled to unit sphere, which requires normalization parameters to project them back to image. We estimate it for the first diffusion model $p_\theta$ when GT is not available and compute them for separate diffusion models $p_\theta^h, p_\theta^o$ from the segmented points. We show in \cref{subsec:exp-ablate-hdm} that it is better than directly predicting from input image. More details in Supp.

\noindent\textbf{Implementation.} We train our diffusion models $p_\theta, p_\theta^h, p_\theta^o$ using the standard loss (~\cref{eq:diffusion-loss}) and segmentation model $g_\phi$ using L2 distance between predicted and ground truth binary labels. See Supp. for more implementation details. %

\section{Experiments}

In this section, we first describe our data generation and then evaluate the proposed \dataName{} data and \modelName{} for reconstruction. Please refer to supp. for implementation details. 

\noindent\textbf{Data generation.} We leverage the BEHAVE~\cite{bhatnagar22behave}, InterCap~\cite{huang2022intercap}, ShapeNet~\cite{shapenet2015}, Objaverse~\cite{objaverse}, ABO~\cite{collins2022abo} and MGN~\cite{bhatnagar2019mgn} dataset to generate our synthetic data \dataName{}. BEHAVE and InterCap capture multi-view images of humans interacting with 20 and 10 different objects respectively. ShapeNet~\cite{shapenet2015}, Objaverse~\cite{objaverse} and ABO~\cite{collins2022abo} provide 3D object models as meshes paired with textures. The objects from ShapeNet and ABO are aligned in canonical space while objects from Objaverse are not aligned. MGN~\cite{bhatnagar2019mgn} consists of 100 human scans paired with SMPL-D registration that allows reposing scans while preserving clothing deformation. 

Following the same split from \cite{xie2023vistracker}, we randomly sample from 380k interactions in BEHAVE and InterCap training set, 21k different shapes in ShapeNet, ABO and Objaverse, and 100 different human shapes and textures in MGN. In total, we generate $\sim$1.1million training images. Please see supplementary for more data distribution details.%

\noindent\textbf{Evaluation metric.}  We evaluate the reconstruction performance using the F-score based on Chamfer distance between point clouds, which is more suitable for measuring the shape accuracy~\cite{what3d_cvpr19}. We compute F-score with a threshold of 0.01m~\cite{melaskyriazi2023pc2} and report the error for human, object and combined point clouds separately, as typically done in interaction reconstruction methods~\cite{xie2023vistracker, xie22chore}.

\subsection{Reconstruction on BEHAVE and InterCap}
We compare our method with CHORE~\cite{xie22chore} and $\text{PC}^2$\cite{melaskyriazi2023pc2} on BEHAVE\cite{bhatnagar22behave} and InterCap~\cite{huang2022intercap} test set in \cref{table:main-results} and \cref{fig:results-comparison}. We train CHORE and \PCTwo{} on the training set of BEHAVE and InterCap. Our \modelName{} is trained on our synthetic \dataName{} with or without BEHAVE and InterCap training set. We also report per-category accuracy in supplementary.  

CHORE is designed for interaction reconstruction and requires \emph{known} object templates. \PCTwo{} is a general shape reconstruction method without templates but it does not separate human and object hence cannot reason the semantics of interaction. Our method trained only on our synthetic \dataName{} dataset performs on par with CHORE which already knows the template and \PCTwo{} which already sees the object shapes. After training our \modelName{} on both our \dataName{} and real data,  our method significantly outperforms baselines. 
\begin{table}[ht]
    \centering
    \small
    \begin{tabular}{c| c | c c c }
    \toprule[1.5pt]
         & Method & Human$\uparrow$ & Object$\uparrow$ & Comb.$\uparrow$  \\
         \hline
         \parbox[t]{1mm}{\multirow{4}{*}{\rotatebox[origin=c]{90}{BEHAVE}}} & CHORE$^\dagger$&0.3454& 0.4258 & 0.3966 \\
         & \PCTwo{}$^\ddagger$ & \text{\sffamily X} & \text{\sffamily X} & 0.4231 \\
         & Ours$^\ddagger$  &{\bf0.3925}& {\bf 0.5049} & {\bf 0.4604} \\
        \cline{2-5}
        & Ours synth. only$^\ddagger$ & 0.3477 & 0.4351 & 0.4110 \\
        \midrule
        \parbox[t]{1mm}{\multirow{4}{*}{\rotatebox[origin=c]{90}{InterCap}}} & CHORE$^\dagger$ & 0.4064 & 0.5135 & 0.4687 \\
         & \PCTwo{}$^\ddagger$ & \text{\sffamily X} & \text{\sffamily X} & 0.5057  \\
         & Ours$^\ddagger$ & {\bf0.4399}& {\bf 0.6072 } & {\bf 0.5344} \\
        \cline{2-5}
        & Ours synth. only$^\ddagger$ & 0.3851 & 0.4928 & 0.4530 \\
    \bottomrule[1.5pt]
    \end{tabular}
    \caption{\textbf{Reconstruction results} (F-sc.@0.01m) on BEHAVE~\cite{bhatnagar22behave} and InterCap~\cite{huang2022intercap}. $^\dagger$ denotes methods with template meshes while $^\ddagger$ denotes template-free methods. CHORE~\cite{xie22chore} requires known object templates and is prone to noisy pose predictions. \PCTwo{}~\cite{melaskyriazi2023pc2} does not require templates but cannot predict semantics of human-object and the prediction is inaccurate.  Our method separates human and object, does not require any templates and outperforms \PCTwo{} and CHORE. Training only on our synthetic \dataName{} data performs on par with CHORE even it has never seen the objects.
    }
    \label{table:main-results}
\end{table}

\subsection{Contribution of our \dataName{} and \modelName{} }
We propose \dataName{} for interaction data generation and \modelName{} for interaction reconstruction. To decouple the contribution of our data and method, we compare our method against \PCTwo{}~\cite{melaskyriazi2023pc2} trained on BEHAVE~\cite{bhatnagar22behave} only (\cref{table:data-vs-method} a-b) and BEHAVE + our \dataName{} dataset (\cref{table:data-vs-method} c-d). The methods are evaluated on the BEHAVE test set. It can be seen that both our proposed data and model are important to obtain the most accurate reconstruction. We also report the model performance vs. data amount in supplementary. 

\begin{table}[]
    \centering
    \small
    \begin{tabular}{c|c c c}
         Method & Human & Object & Combined \\
         \hline
          a. \PCTwo & \text{\sffamily X} & \text{\sffamily X} & 0.4231  \\
          b. Our \modelName{} & 0.3605 & 0.4575 & 0.4214 \\
         \hline
         c. \PCTwo{} + our \dataName{} & \text{\sffamily X} & \text{\sffamily X} & 0.4486  \\
        d. Our \modelName{} + \dataName{} & {\bf 0.3925 } & {\bf 0.5049} & {\bf 0.4604} \\
    \end{tabular}
    \caption{\textbf{Decoupling the contribution} of our \dataName{} dataset and reconstruction method. Our \dataName{} dataset significantly boosts performance of both \PCTwo{} (c) and our method (d) compared to training on BEHAVE only (a-b).
    Both our \dataName{} and \modelName{} model are important to achieve the best result.}
    \label{table:data-vs-method}
\end{table}

\begin{figure*}[t]
    \centering
    \includegraphics[width=1.0\linewidth]{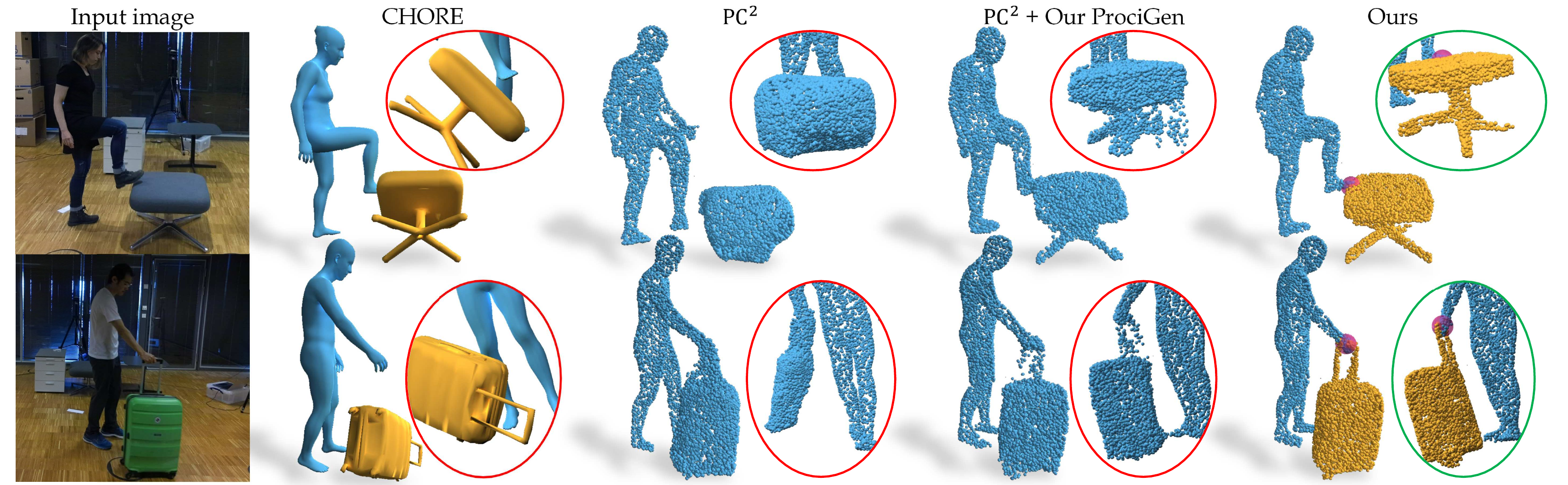}
    \caption{\textbf{Generalization results to InterCap~\cite{huang2022intercap} dataset.} Note that all object instances are unseen during training time. CHORE~\cite{xie22chore} predicts template specific object pose hence cannot generalize to new object instances. \PCTwo{}\cite{melaskyriazi2023pc2} does not rely on template but its generalization ability is constrained by limited shape variations from BEHAVE~\cite{bhatnagar22behave}. Training \PCTwo{} on our \dataName{} improves its generalization but the predicted point clouds are still noisy. Our method is able to generalize and predicts human and object with high shape fidelity. }
    \label{fig:generalization-comp}
\end{figure*}

\subsection{Generalization Performance}
\label{sec:exp-generalization}
Our \dataName{} dataset allows training shape reconstruction methods to generalize to unseen object instances. To evaluate this, we train CHORE\cite{xie22chore}, \PCTwo{}\cite{melaskyriazi2023pc2} and our \modelName{} model on BEHAVE\cite{bhatnagar22behave} and our proposed dataset respectively. We then evaluate them on unseen objects of the same categories from InterCap~\cite{huang2022intercap} in \Cref{table:generalization}. CHORE requires a template to predict 6D pose, which makes it difficult to train on our synthetic dataset with more than 21k different shapes. We hence only train CHORE on BEHAVE dataset.  %

Methods trained on BEHAVE have limited generalization to InterCap (\cref{table:generalization}a-c). An alternative to our \dataName{} is to randomly scale and shift the objects from BEHAVE and render new images, which only slightly improves generalization (\cref{table:generalization}d). In contrast, our \dataName{} significantly boosts the generalization performance (\cref{table:generalization}e-f). 

\begin{figure*}[t]
    \centering
    \includegraphics[width=1.0\linewidth]{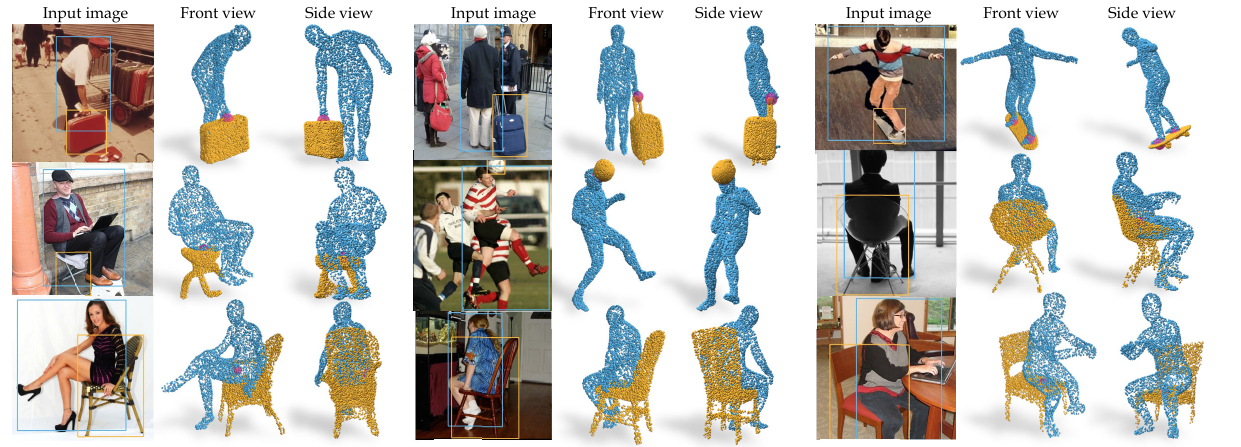}
    \caption{\textbf{Testing our method on COCO~\cite{coco-dataset} dataset.} Human and object to be reconstructed are highlighted with blue and yellow box respectively. Our method generalizes to diverse objects from in the wild images \emph{without any shape templates}.}
    \label{fig:results-generalization}
\end{figure*}

Some qualitative results are shown in \cref{fig:generalization-comp}. Our method reconstructs human and object with high shape fidelity.  We also show the generalization results to COCO dataset~\cite{coco-dataset} in \cref{fig:results-generalization}. Our method trained \emph{only} on our \dataName{} data generalizes well to in-the-wild images with large object shape variations. See Supp. for more generalization examples. 

\begin{table}[]
    \centering
    \small
    \begin{tabular}{l|c c c }
         Method & Human $\uparrow$ & Object$\uparrow$ & Combine$\uparrow$  \\
         \hline
         a. CHORE & 0.2263 & 0.1924 & 0.2176 \\
         b. \PCTwo{} & \text{\sffamily X} & \text{\sffamily X} & 0.2327 \\
         c. Our \modelName & 0.2389 & 0.1592 & 0.2127 \\ 
         d. Our \modelName + augm. & 0.3076 & 0.2089 & 0.2680 \\ 
         e. \PCTwo{} + Our \dataName & \text{\sffamily X} & \text{\sffamily X} & 0.3843 \\
         f. Our \modelName + \dataName & \bf 0.3502 & \bf 0.4233 & \bf 0.3976 \\ 
    \end{tabular}
    \caption{\textbf{Generalization performance} of methods trained on BEHAVE~\cite{bhatnagar22behave} (a-c), BEHAVE + random augmentation (d) and our \dataName{} (e-f),  evaluated on unseen objects from InterCap (F-score@0.01m). CHORE predicts template-specific 6D poses hence does not work on unseen objects from InterCap. \PCTwo{} (b) and our method (c) do not require templates but are constrained by the limited shape variations from BEHAVE. Adding random shape augmentation on BEHAVE objects (d) slightly improves generalization but is still suboptimal. 
    With our proposed \dataName{} dataset, both \PCTwo{} and our method can generalize to InterCap and our method achieves better accuracy. 
    }
    \label{table:generalization}
\end{table}

\subsection{Ablating the Hierarchical Diffusion Model}\label{subsec:exp-ablate-hdm}
Our \modelName{} predicts interaction semantics and better shapes. In \cref{table:ablate-hdm}, we ablate alternative designs to our method on the 824 chair images from BEHAVE test set~\cite{bhatnagar22behave} due to resource limit. All methods are trained on our \dataName{} dataset. 

The human-object segmentation allows us to compute the contacts and manipulate human and object separately. An alternative is projecting the predicted points to 2D image and segment points based on the masks. Due to occlusion and complex interaction, this segmentation is inaccurate, as reflected in the large human and object errors in \cref{table:ablate-hdm}a. The model that predicts human, object and segmentation with a single model (\cref{table:ablate-hdm} b) also does not work as it is difficult to learn high-fidelity interaction shapes. 

Another alternative to our first joint diffusion model is to use a network that predicts translation and scale directly from input image and then use them to combine predictions from two separate models. However, such a global prediction does not model interaction with local fine-level details hence the performance is subpar(\cref{table:ablate-hdm}b). Our cross attention module also improves the performance (\cref{table:ablate-hdm}d). 
\begin{table}[]
    \centering
    \small
    \begin{tabular}{l|c c c }
         Method (with our \dataName{}) & Hum.$\uparrow$ & Obj.$\uparrow$ & Comb.$\uparrow$  \\
         \hline
         a. \PCTwo{} + projected segm. & 0.2961 & 0.3436 & 0.3776 \\
         b. Single model + segm. & 0.3349 & 0.3638 & 0.3743  \\
         c. Direct pred. + sep. models &  0.2809 & 0.3487 & 0.3380 \\
         d. Ours w/o cross attention & 0.3387 & 0.3806 & 0.3807 \\ 
         e. Our full model & \bf 0.3433  & \bf 0.3916 & \bf 0.3875 \\
    \end{tabular}
    \caption{Ablating alternative methods to our \modelName{} (F-score@0.01m). Projecting \PCTwo{} predictions to 2D masks to obtain segmentation (a) is inaccurate and single stage diffusion model (b) cannot learn high-fidelity shapes for both human and object. Combining predictions from separate human and object models using direct translation prediction from images (c) also does not work as it cannot learn fine-grained interactions. Our hierarchical design together with our cross attention module achieves the best result. 
    }
    \label{table:ablate-hdm}
\end{table}

\section{Conclusion}\label{sec:conclusion}
In this paper, we proposed a procedural generation method to synthesize interaction datasets with diverse human and object shapes. This method allows us to generate 1M+ images paired with clean 3D ground truth and train large image-conditioned diffusion models for reconstruction, without relying on any shape templates. To learn accurate shape space for human and object, we introduce a hierarchical diffusion model that learns both the joint interaction and high fidelity human and object shape subspaces. 

We train our method with the proposed synthetic dataset and evaluate it on BEHAVE and InterCap datasets. Results show that our method significantly outperforms CHORE which requires template meshes and \PCTwo{} which does not reason interaction semantics. Ablation studies also show that our synthetic dataset is important to boost the performance and generalization ability of both \PCTwo{} and our model. Our method generalizes well to real images from COCO that have diverse object geometries, which is a promising step toward real in-the-wild reconstruction. Our code and data are released to promote future research. 

\noindent
{\footnotesize
\textbf{Acknowledgements.} We thank RVH group members \cite{rvh_grp}, especially Yuxuan Xue, for their helpful discussions. This work is funded by the Deutsche Forschungsgemeinschaft (DFG, German Research Foundation) - 409792180 (Emmy Noether Programme,
project: Real Virtual Humans), and German Federal Ministry of Education and Research (BMBF): Tübingen AI Center, FKZ: 01IS18039A, and Amazon-MPI science hub. Gerard Pons-Moll is a Professor at the University of Tübingen endowed by the Carl Zeiss Foundation, at the Department of Computer Science and a member of the Machine Learning Cluster of Excellence, EXC number 2064/1 – Project number 390727645.
}

{
    \small
    \bibliographystyle{ieeenat_fullname}
    \bibliography{main}
}

\clearpage
\setcounter{page}{1}
\maketitlesupplementary

In this supplementary, we first discuss in more detail about our implementation for the \dataName{} and \modelName{} in \cref{sec:implementation}. We also present the statistics of our generated \dataName{} dataset. We then show more results and experimental analysis of our method in \cref{sec:supp-exp}. We conclude with a discussion of limitations and future works. 
\appendix
\section{Implementation Details}
We describe in more details of the implementation of our \dataName{} and \modelName{}. Our code for both data generation and reconstruction will be made publicly available. 
\label{sec:implementation}
\subsection{\dataName{} Data Generation}
\textbf{Correspondence estimation.} We use the implementation from ART~\cite{zhou2022art} for our autoencoder, which uses PointNet~\cite{Qi2016pointnet} as encoder and 3-layer MLPs as decoder. We sample 8000 points from the mesh surface and train the network with bidirectional Chamfer distance. To ensure reconstruction quality, we overfit one network per category. Each model is trained for 5000 epochs. We report an average reconstruction error of around 7mm for our autoencoders, which indicates highly accurate reconstructions. 

\noindent\textbf{Contact transfer and optimization}. We use a threshold of $\sigma=2$cm to find points that are in contact. The loss weights for our contact based loss optimization are: $\lambda_c=400, \lambda_n=6.25, \lambda_\text{colli}=9, \lambda_\text{init}=6.25\cdot 10^4$.

\noindent\textbf{Rendering}. We use blender to render our synthesized human-object interactions. We choose one set of 4 camera configurations from BEHAVE~\cite{bhatnagar22behave} and another set of 6 camera configurations from InterCap~\cite{huang2022intercap}. For each synthesized interaction, we additionally add small random global rotation and translation to have variations of camera viewpoints. We render the interactions with an empty background since our network also takes images with background masked out as input. We add lights at fixed locations with random light intensities. Our blender scene and rendering code will also be made publicly available. 

\begin{figure*}[t]
    \centering
    \includegraphics[width=\linewidth]{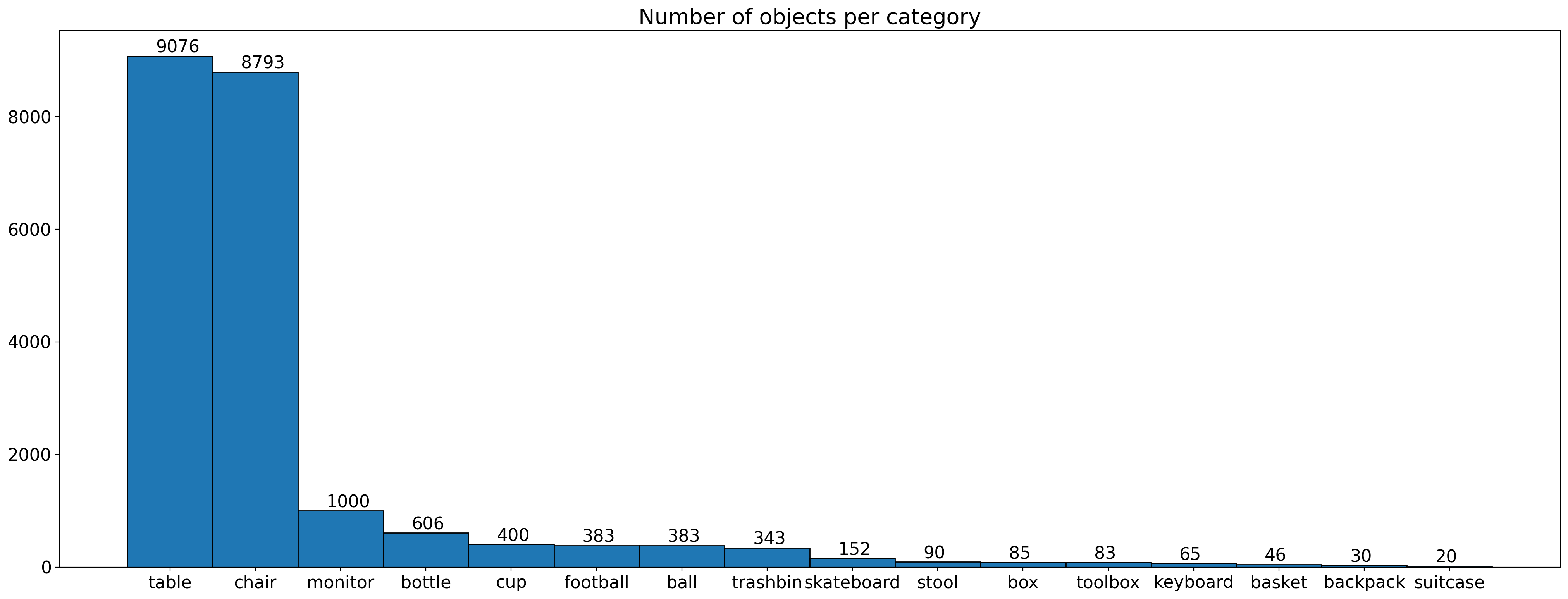}
    \caption{The number of objects per category we used to generate our \dataName{} dataset. It can be seen that the shape variations are dominated by tables and chairs, which are also the categories with the most complex shapes.}
    \label{fig:supp-distribution-shapes}
\end{figure*}

\begin{figure*}[t]
    \centering
    \includegraphics[width=\linewidth]{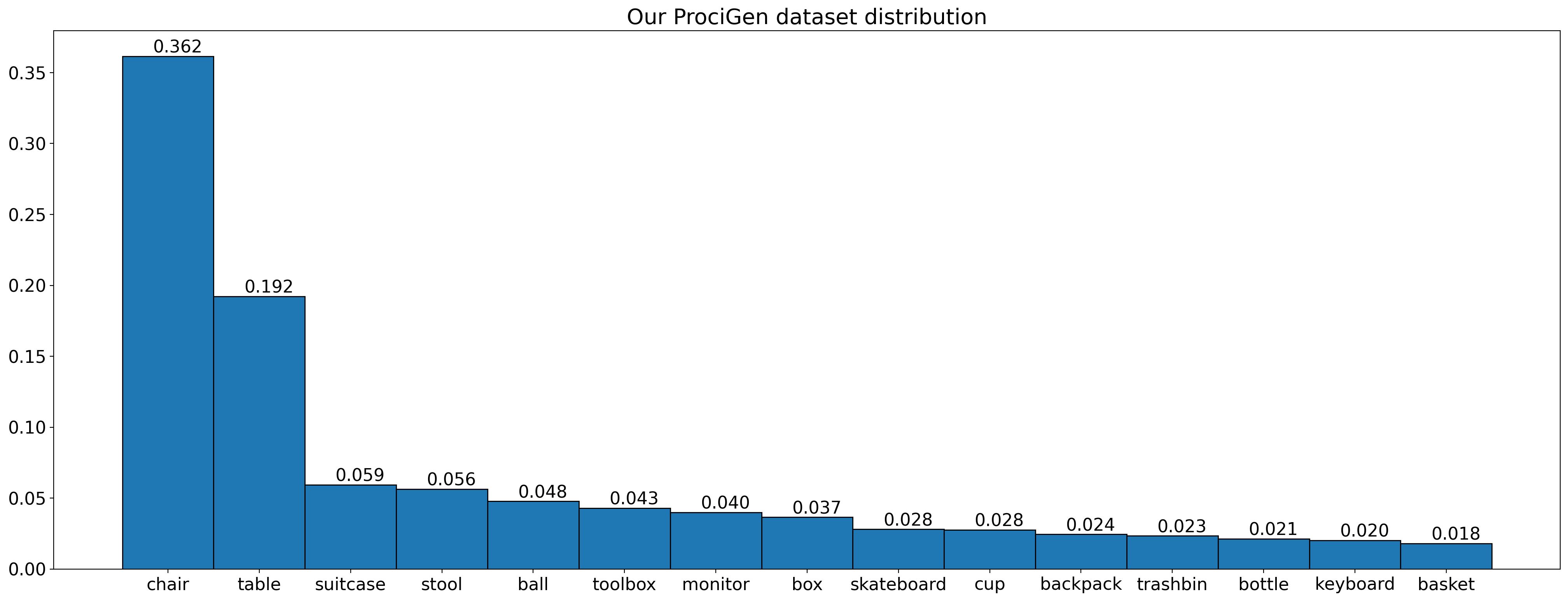}
    \caption{Distribution of interactions per object category. Our dataset features interaction data with very diverse object shapes, which is not possible via real data capture.}
    \label{fig:supp-distribution-interactions}
\end{figure*}

\subsection{\modelName{}: Hierarachical Diffusion Model}
We use the modified Point Voxel CNN from ~\cite{Zhou_2021_ICCV_PVD} as the network for our joint diffusion $p_\theta$, segmentation $g_\phi$, and separate diffusion models $p_\theta^h, p_\theta^o$. The input images are cropped and resized to $224\times 224$.  The joint diffusion model diffuses in total 16384 points while the separate models diffuse 8196 points each. We use the MAE~\cite{he_mae_2022} as the image feature encoder. We additionally stack the human and object masks as well as distance transform as additional image features, same as \PCTwo{}~\cite{melaskyriazi2023pc2}. 
We train our diffusion models for a total of 500000 steps with batch size 20. We use a linear scheduler without warm-up for the forward diffusion process, in which beta increases from $1\cdot 10^{-5}$ to $8\cdot 10^{-3}$. For the network optimization, we use AdamW optimizer with linear learning rate decay starting from $3\cdot 10^{-4}$ and decreasing to 0 during the course of training. The diffusion models are trained with the standard diffusion training scheme~\cite{ho2020ddpm}. To train the segmentation model, we add small Gaussian noise to the GT point clouds and project them to obtain image features. The loss is then computed between the prediction and recomputed GT labels on the points with noise. To speed up training, we train stage 1 ($g_\phi, p_\theta$) and stage 2 ($p_\theta^h, p_\theta^o$) models separately. For each stage, it takes around 4 days to train on a machine with 4 A40 GPUs.

\textbf{Camera estimation.} Recall from \cref{subsec:hierarchical-diffusion} that a camera translation is required to project the normalized point clouds back to the image. This needs to be estimated from input when GT camera pose is not available, especially for generalization to diverse datasets. The camera translation consists of three unknowns, which requires at least two point pairs of 3D location and 2D-pixel coordinates. We empirically choose the Gaussian point center and one edge of the point cloud. The idea is to have the initial Gaussian point clouds cover the 2D human object interaction region and the 3D center is projected to 2D crop center. 

Formally, let $\vect{p}_c=(c_x, c_y)$ be the center coordinate of the 2D interaction region, $w$ be the width of the 2D interaction square crop, $\vect{p}_e=(\sigma, 0, z)$ be a 3D point near the edge of the Gaussian sphere with unknown depth $z$. Given camera projection matrix $\mat{K}\in \mathbb{R}^{3\times 3}$ and translation vector $\vect{t}_c$, we define the following equations:
\begin{equation}
   \mat{K}\vect{t}_c = \vect{p}_c;\ \ \mat{K}(\vect{p}_e+\vect{t}_c) =\vect{p}^\text{2D}_e
\end{equation}
The first equation projects origin to $\vect{p}_c$ and the second equation projects $\vect{p}_e$ to the middle right of the 2D crop $\vect{p}^\text{2D}_e=(c_x+w/2, c_y)$. This is a linear system of four equations with four unknowns (camera translation and depth $z$), leading to a unique solution for the translation $\vect{t}_c$. We empirically set $\sigma$ to different values for different categories based on the estimation error on the BEHAVE training set. Furthermore, we compute $\vect{p}_c$ as the centroid of all 2D points inside the human and object masks. From \cref{fig:results-generalization}, \cref{fig:supp-sysu}, \cref{fig:supp-ntu}, \cref{fig:supp-coco01}, \cref{fig:supp-coco02} and , \cref{fig:supp-coco03}, it can be seen that our method can reconstruct human and object well on different datasets using our estimated translation. 

\subsection{Dataset statistics}
We generate our \dataName{} dataset based on interactions from BEHAVE~\cite{bhatnagar22behave} and InterCap~\cite{huang2022intercap}, human scans from MGN~\cite{bhatnagar2019mgn}, object shapes from ShapeNet~\cite{shapenet2015}, Objaverse~\cite{objaverse} and ABO~\cite{collins2022abo}. When generating our data, we mainly consider the variation of object shapes and interaction poses while the object sizes remain the same. We also try to avoid large imbalances among objects. Therefore, chairs and tables are two dominant categories as they have the most geometry and interaction pose variations (\cref{fig:supp-distribution-shapes}, \cref{fig:supp-distribution-interactions}). Other categories have similar amounts of synthetic data as they have similar amounts of interaction poses. The difference comes from failures in joint optimization due to irregular mesh.

In total we rendered 1.1M interaction images with 21555 different object shapes. The distribution for object shapes and interactions per category are shown in \cref{fig:supp-distribution-shapes} and \cref{fig:supp-distribution-interactions}. Our dataset has very diverse object shapes, especially for chairs and tables whose geometry also varies a lot in reality. Our procedural generation method is a scalable solution and it allows for generating large-scale interaction datasets with great amount of variations which is not obtainable via capturing real data.

\section{Additional Experiments and Results}\label{sec:supp-exp}
\subsection{Per-category reconstruction accuracy}
We report the accuracy of each category in \cref{tab:category-fscore}. Our method consistently outperforms baselines in almost all categories. 
While the improvements in numbers look small, the visual difference is quite significant, as shown in the paper \cref{fig:results-comparison}, \cref{fig:results-generalization}. 
\begin{table}[t]
    \centering
    \scriptsize
    \begin{tabular}{p{1.3cm} |>{\centering\arraybackslash}p{0.4cm} >{\centering\arraybackslash}p{0.4cm} >{\centering\arraybackslash}p{0.5cm} |>{\centering\arraybackslash}p{0.4cm} |>{\centering\arraybackslash} p{0.4cm} >{\centering\arraybackslash}p{0.4cm} >{\centering\arraybackslash}p{0.5cm}}
    \toprule
         \multirow{2}{*}{Method} & \multicolumn{3}{c|}{CHORE $\uparrow$} & \multicolumn{1}{c|}{PC2 $\uparrow$}& \multicolumn{3}{c}{ Ours $\uparrow$} \\
         & Hum. & Obj. & Comb. & Comb. & Hum. & Obj. & Comb. \\
         \hline
         Chair & 0.373 & 0.491 & 0.443&0.407 & {\bf 0.384 } & {\bf 0.521 } & {\bf 0.463 } \\
         Ball & 0.330 & 0.388 & 0.374&0.424 & {\bf 0.395 } & {\bf 0.517 } & {\bf 0.471 } \\
    Backpack & {\bf 0.399} & {\bf 0.509} & {\bf 0.469} & 0.436 & 0.397 & 0.457 & 0.444 \\
    Table & 0.304 & 0.455 & 0.389&0.470 & {\bf 0.379 } & {\bf 0.642 } & {\bf 0.517 } \\
    Basket & 0.301 & 0.266 & 0.292&{\bf 0.381} & {\bf 0.412 } & {\bf 0.297 } & 0.364 \\
    Box & 0.352 & 0.347 & 0.362&0.409 & {\bf 0.414 } & {\bf 0.401 } & {\bf 0.424 } \\
    Keyboard & 0.335 & 0.412 & 0.383&0.450 & {\bf 0.353 } & {\bf 0.606 } & {\bf 0.493 } \\
    Monitor & 0.358 & {\bf 0.412} & {\bf 0.395}&0.377 & {\bf 0.368 } & 0.348 & 0.370 \\
    Suitcase & 0.400 & 0.477 & 0.443&0.404 & {\bf 0.431 } & {\bf 0.484 } & {\bf 0.462 } \\
    Stool & 0.351 & 0.479 & 0.424&0.443 & {\bf 0.394 } & {\bf 0.543 } & {\bf 0.479 } \\
    Toolbox & 0.281 & 0.330 & 0.306&0.398 & {\bf 0.373 } & {\bf 0.400 } & {\bf 0.403 } \\
    Trashbin & 0.376 & 0.402 & 0.398&0.387 & {\bf 0.407 } & {\bf 0.414 } & {\bf 0.422 } \\
    \hline
    BEHAVE all & 0.345&0.426&0.397&0.423& {\bf 0.392 } & {\bf 0.498 } &{\bf 0.457 } \\
    \midrule
    Chair & {\bf 0.389} & 0.468 & 0.433&0.470 & 0.384 & {\bf 0.604 } & {\bf 0.500 } \\
    Cup & 0.412 & 0.538 & 0.510&0.566 & {\bf 0.487 } & {\bf 0.601 } & {\bf 0.578 } \\
    Skateboard & {\bf 0.520} & 0.684 & 0.612&0.578 & 0.491 & {\bf 0.739 } & {\bf 0.624 } \\
    Bottle & 0.426 & 0.501 & 0.495&0.592 & {\bf 0.549 } & {\bf 0.582 } & {\bf 0.593 } \\
    \hline
    InterCap all & 0.406&0.513&0.469&0.506& {\bf 0.440 } & {\bf 0.607 } &{\bf 0.534 } \\
    \toprule
    \end{tabular}
    \captionsetup{skip=1pt} %
    \caption{Per-category F-score@0.01m comparison. Note that PC2 cannot separate human-object hence we only report the combined error, and that CHORE requires template meshes. Our method outperforms baselines for almost all categories.}
    \label{tab:category-fscore}
\end{table}

\subsection{Performance vs. amount of data}
\begin{figure}
    \vspace{-8pt}  %
    \centering
    \includegraphics[width=1.0\linewidth]{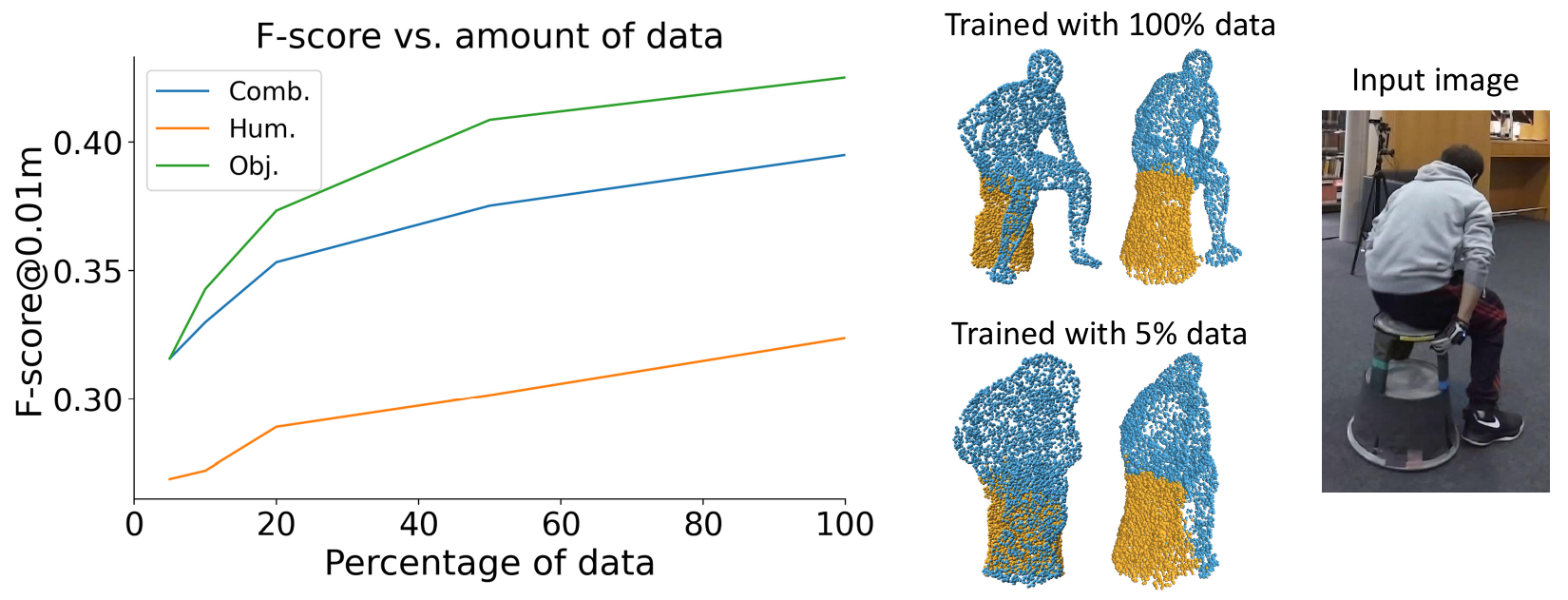}
    \captionsetup{skip=1pt} %
    \caption{Reconstruction performance vs. amount of data. It can be seen that more data leads to better results. %
    }
    \label{fig:fscore-vs-data}
\end{figure}

We show in \Cref{table:data-vs-method} that our data contributes a lot to improve the reconstruction accuracy. To further understand the data contribution, we train our model for the same epochs with different amounts of synthetic data and test on BEHAVE images without fine-tuning. 
The performance vs. data plot is shown in \cref{fig:fscore-vs-data}. More data consistently leads to better performance both quantitatively and qualitatively. 

\subsection{Analysis of $T_0$ for our \modelName{}}
\begin{figure}[t]
    \centering
    \includegraphics[width=\linewidth]{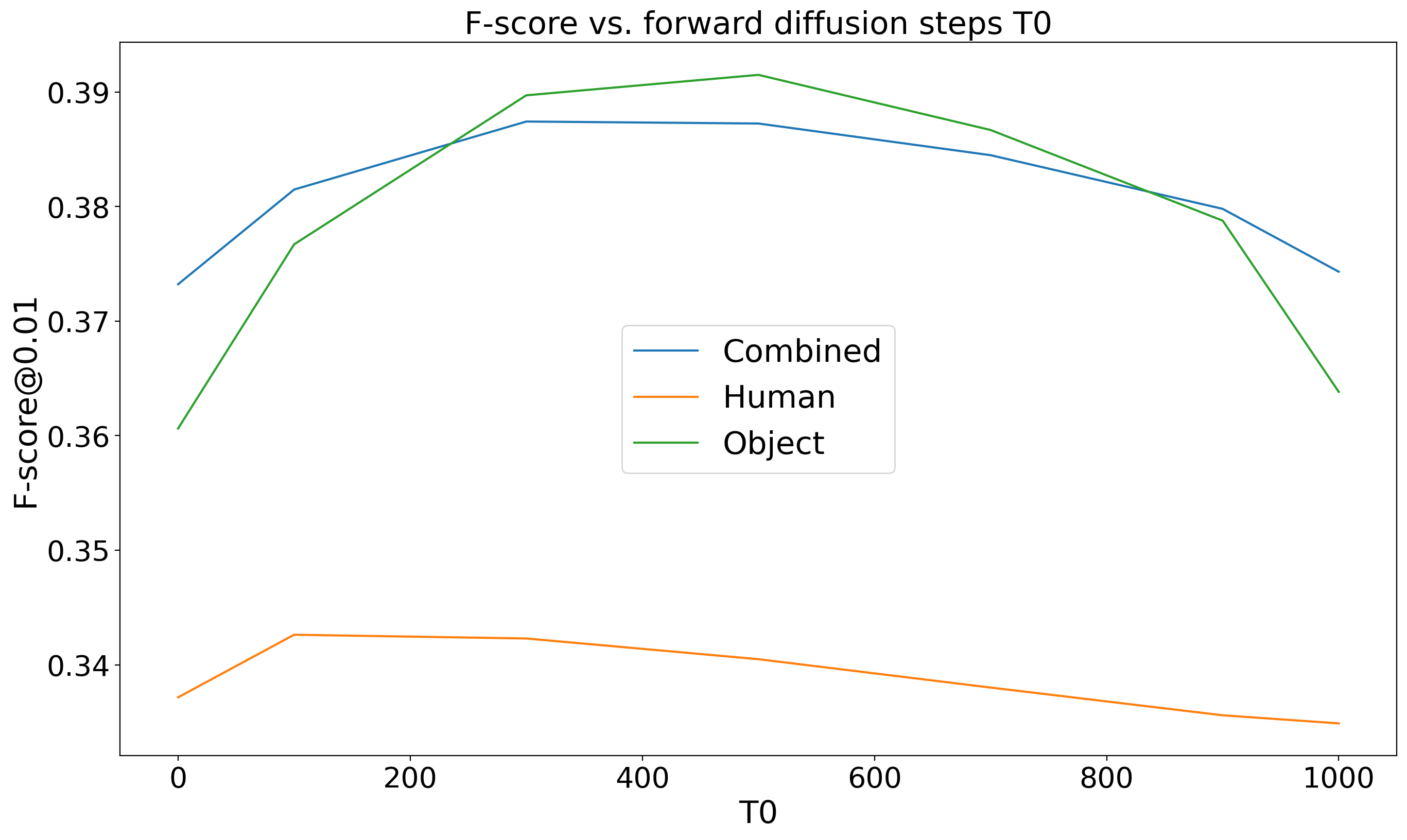}
    \caption{The performance of our method using different intermediate step $T_0$ for the input to our second stage diffusion. Methods are evaluated using F-score@0.01m. At $T_0=500$, we obtain a good balance between human and object performance.}
    \label{fig:fscore-vs-steps}
\end{figure}

In our second stage, we first add noise to the clean predictions from stage one until step $t=T_0$, and then run the reverse diffusion process from $t=T_0$ to $t=0$. We evaluate the performance of our method under different values of $T_0$ in \Cref{fig:fscore-vs-steps}. There is a trade-off for the number of forward steps $T_0$: with a larger $T_0$, less interaction information and noisy details are preserved and the network predicts sharper detail but less faithful to initial prediction and interaction constraints. It can be seen that $T_0=500$ is a good balance between shape fidelity and interaction coherence. 

\subsection{Shape fidelity}\label{subsec:shape-fidelity}
\begin{figure*}
    \centering
    \includegraphics[width=0.98\linewidth]{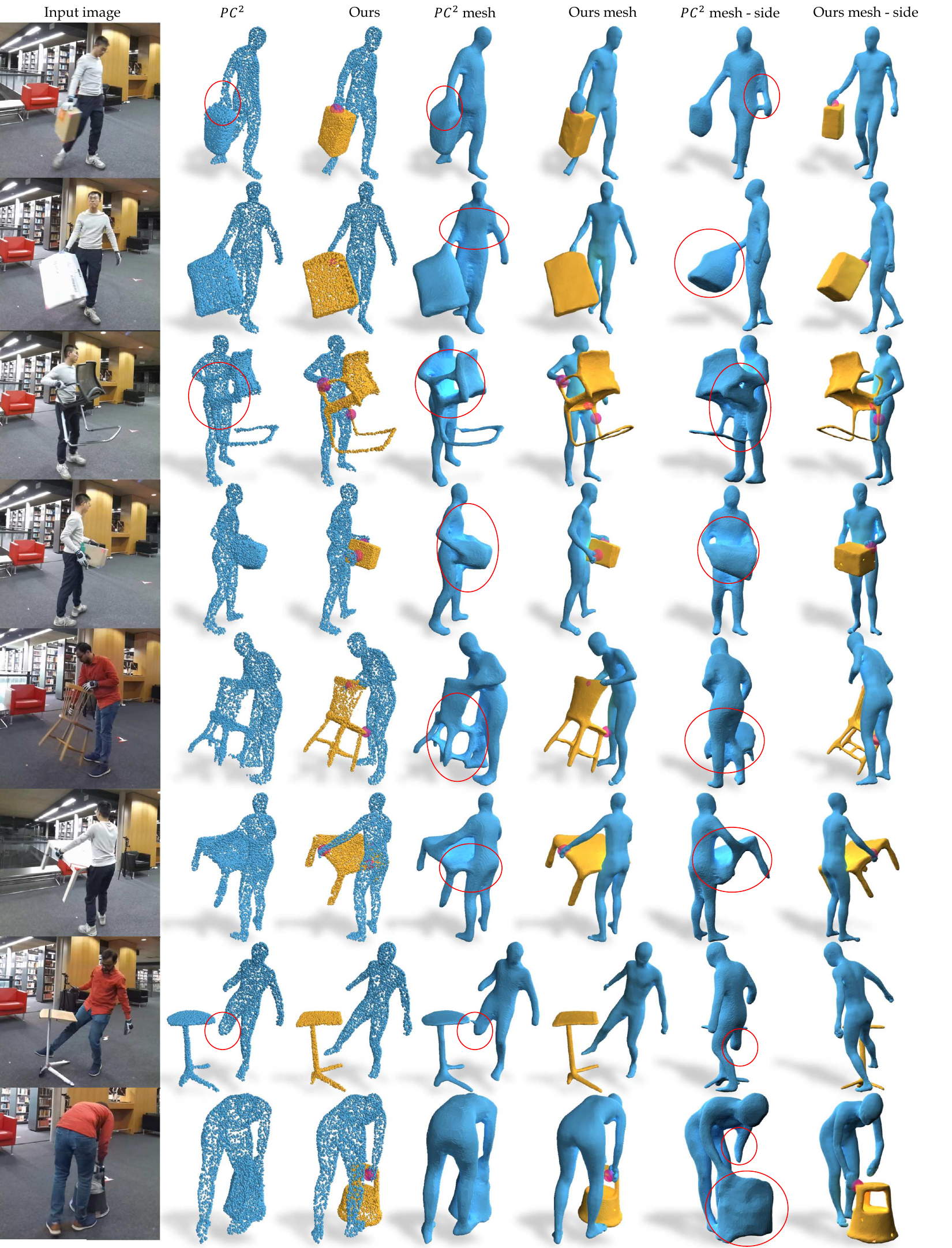}
    \caption{Comparing the shape fidelity of our method with \PCTwo{} on the BEHAVE~\cite{bhatnagar22behave} dataset. \PCTwo{} does not separate human and object and its prediction is noisy, leading to inaccurate meshes. Our method predicts clean point clouds with human object segmentations, allowing us to extract high-quality mesh surfaces. }
    \label{fig:supp-mesh}
\end{figure*}
Our method predicts dense and clean point clouds which are ready for accurate surface extraction. We show in \cref{fig:supp-mesh} that high-quality meshes can be extracted from our predicted point clouds. More specifically, we use screened Poisson surface reconstruction for the human points using normals estimated by MeshLab. For the object, we first use Delaunay triangulation to obtain triangle mesh. We then run fusion-based waterproofing~\cite{Stutz2018mesh_fusion} to obtain a watertight mesh. We also apply Delaunay triangulation and waterproofing to \PCTwo{}~\cite{melaskyriazi2023pc2} predictions and results are shown in \cref{fig:supp-mesh}. It can be seen that \PCTwo{} predictions have missing structure and noisy point clouds, leading to low-quality meshes. In contract, we can extract high-quality meshes directly from our point cloud reconstructions, without any post processing. 

\begin{figure}[t]
    \centering
    \includegraphics[width=\linewidth]{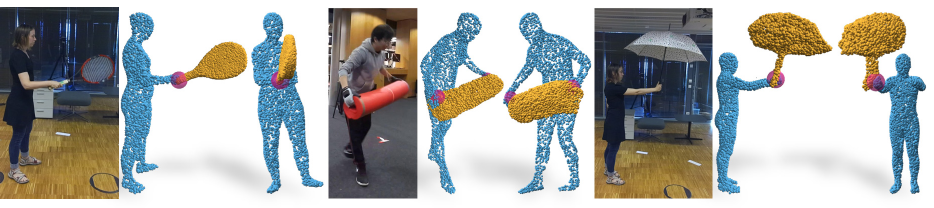}
    \captionsetup{skip=1pt} %
    \caption{Out of distribution generalization. Our method can reconstruct some categories that are \emph{unseen} in training data.}
    \label{fig:ood-generalization}
\end{figure}

\subsection{Interaction semantics}
\begin{figure*}
    \centering
    \includegraphics[width=\linewidth]{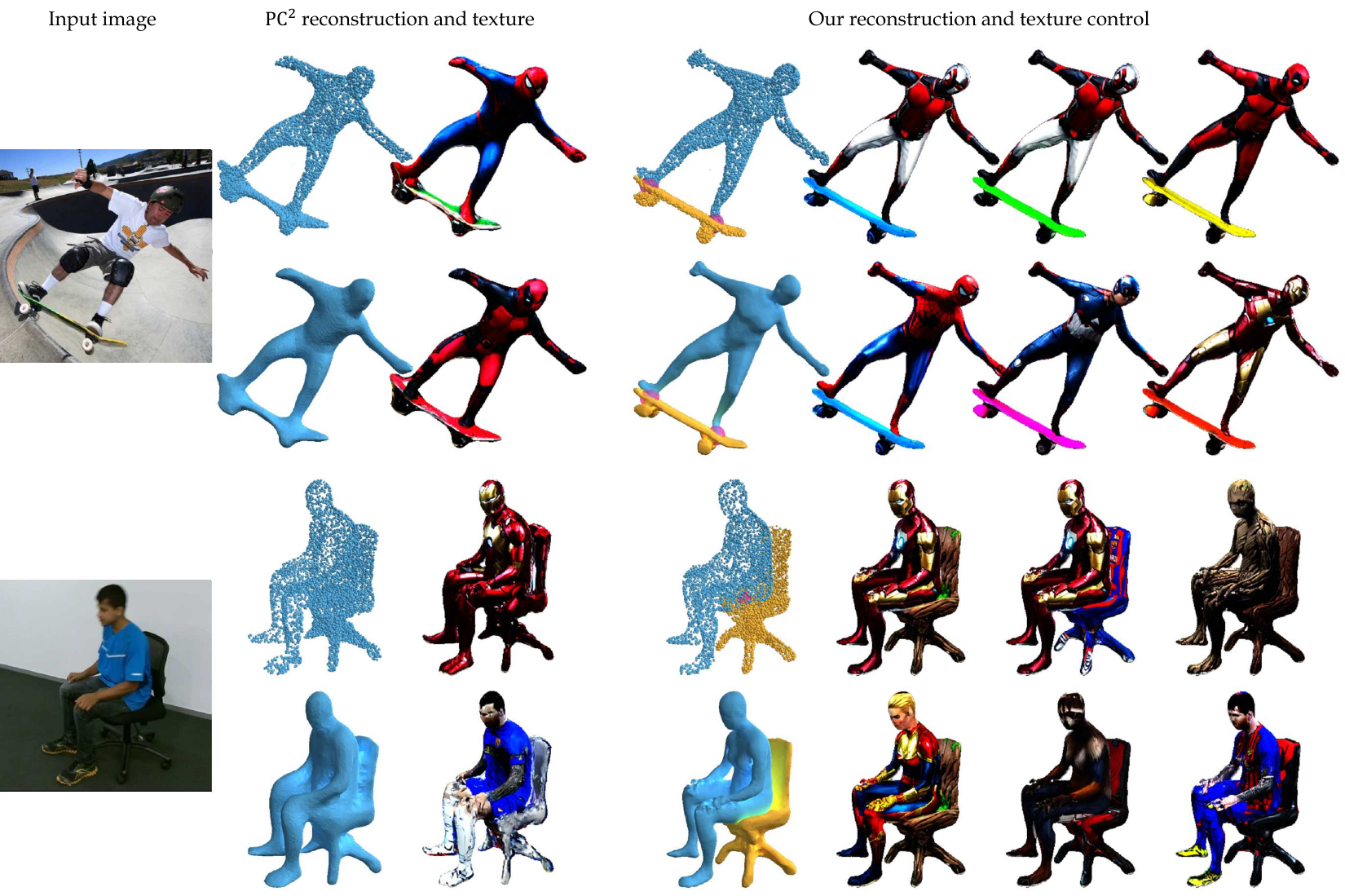}
    \caption{Comparing textures generated for meshes extracted from \PCTwo{}~\cite{melaskyriazi2023pc2} and our predicted point clouds. Textures are obtained using Text2txt~\cite{chen2023text2tex}. \PCTwo{} predicts human and object as one joint point cloud with noisy points, which leads to inaccurate mesh surfaces and it is difficult to generate textures for this combined mesh. It also does not allow changing human and object textures separately. Our method predicts high quality point clouds with segmentation. This enables us to extract high-fidelity mesh, which is important for generating high-quality texture and manipulating human and object differently. }
    \label{fig:supp-texture}
\end{figure*}

Our method predicts the segmentation of human and object, allowing separate manipulation which is important for downstream applications. To demonstrate this, we use Text2txt~\cite{chen2023text2tex} to generate textures for the meshes extracted from \PCTwo{} and our predicted point clouds. Other methods such as Paint-it~\cite{youwang2023paintit} are also applicable here. We show the reconstruction and generated textures in \cref{fig:supp-texture}. It can be seen that \PCTwo{} predictions are noisy and it does not reason human and object separately. This leads to low-quality mesh and generating coherent texture for a combined mesh of human and object is difficult. On the contrary, our method separate human and object while also predicting high quality individual shapes. This allows generating high quality texture and changing textures for human and object differently.  

\begin{figure*}
    \centering
    \includegraphics[width=\linewidth]{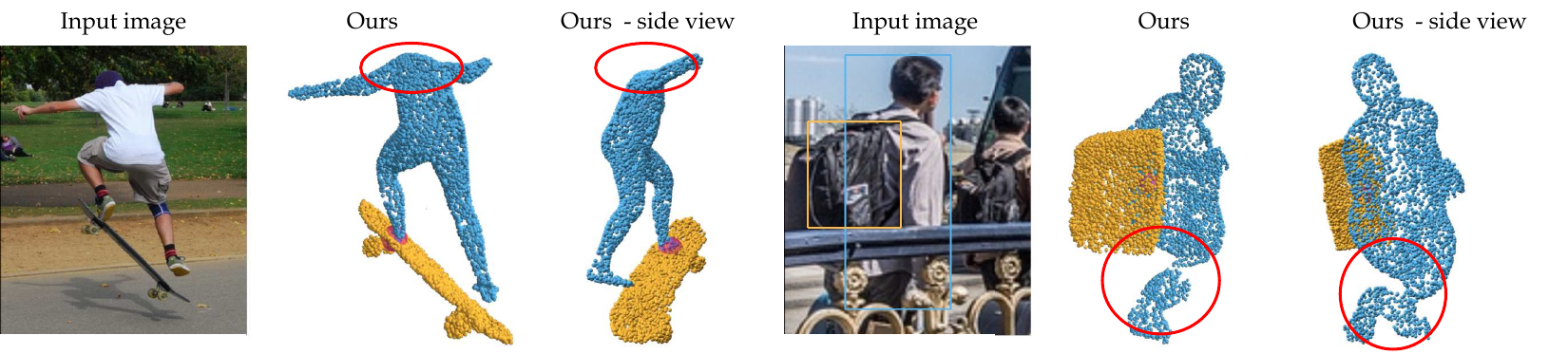}
    \caption{Example failure cases of our method. Our method can fail when large parts of human body are invisible, leading to incoherent human shape reconstructions. Future works can explore human body shape priors to regularize the network predictions.}
    \label{fig:supp-failure}
\end{figure*}

\subsection{More generalization results}
\begin{figure*}
    \centering
    \includegraphics[width=0.95\linewidth]{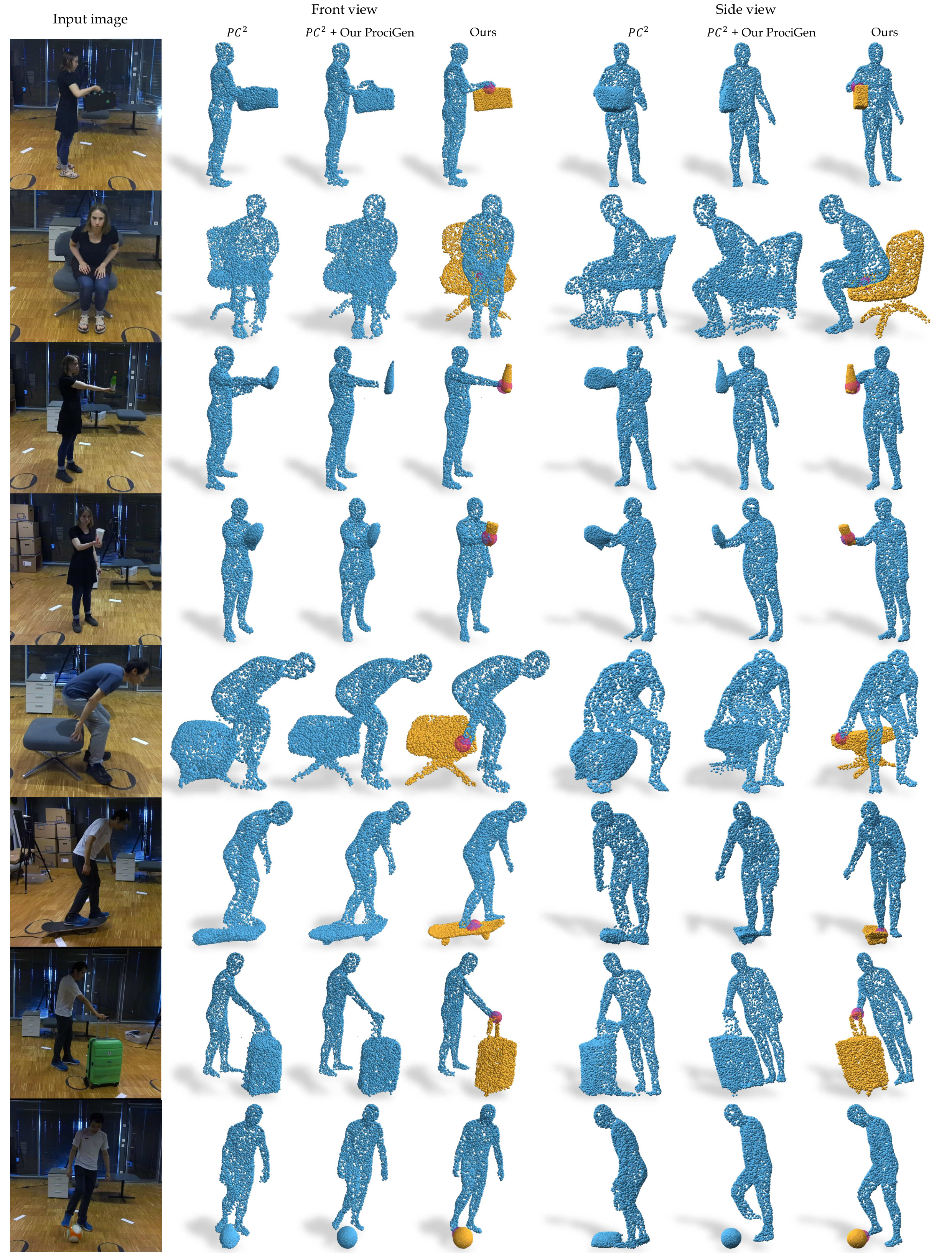}
    \caption{Comparing generalization performance on InterCap~\cite{huang2022intercap}. All objects are unseen during training time. \PCTwo{} trained only on BEHAVE~\cite{bhatnagar22behave} has limited generalization ability. Training \PCTwo{} with our \dataName{} improves generalization but it still cannot reason human and object separately and the predicted points are noisy. Our method trained only on our \dataName{} generalizes well to InterCap objects even they are completely unseen.}
    \label{fig:supp-intercap}
\end{figure*}
We show more generalization comparison on the InterCap~\cite{huang2022intercap} dataset in \cref{fig:supp-intercap}. Note that all objects from InterCap are unseen during training time. It can be seen that \PCTwo{} trained on BEHAVE~\cite{bhatnagar22behave} only cannot generalize to objects from InterCap. Training \PCTwo{} with our \dataName{} dataset allows better generalization ability but its shape prediction is still noisy. Furthermore, \PCTwo{} cannot segment human and object, which is important to reason the interaction semantics and manipulate them separately. Our method generalizes well to InterCap and reconstructs high quality shapes with interaction semantics. 
 
\begin{figure*}
    \centering
    \includegraphics[width=\linewidth]{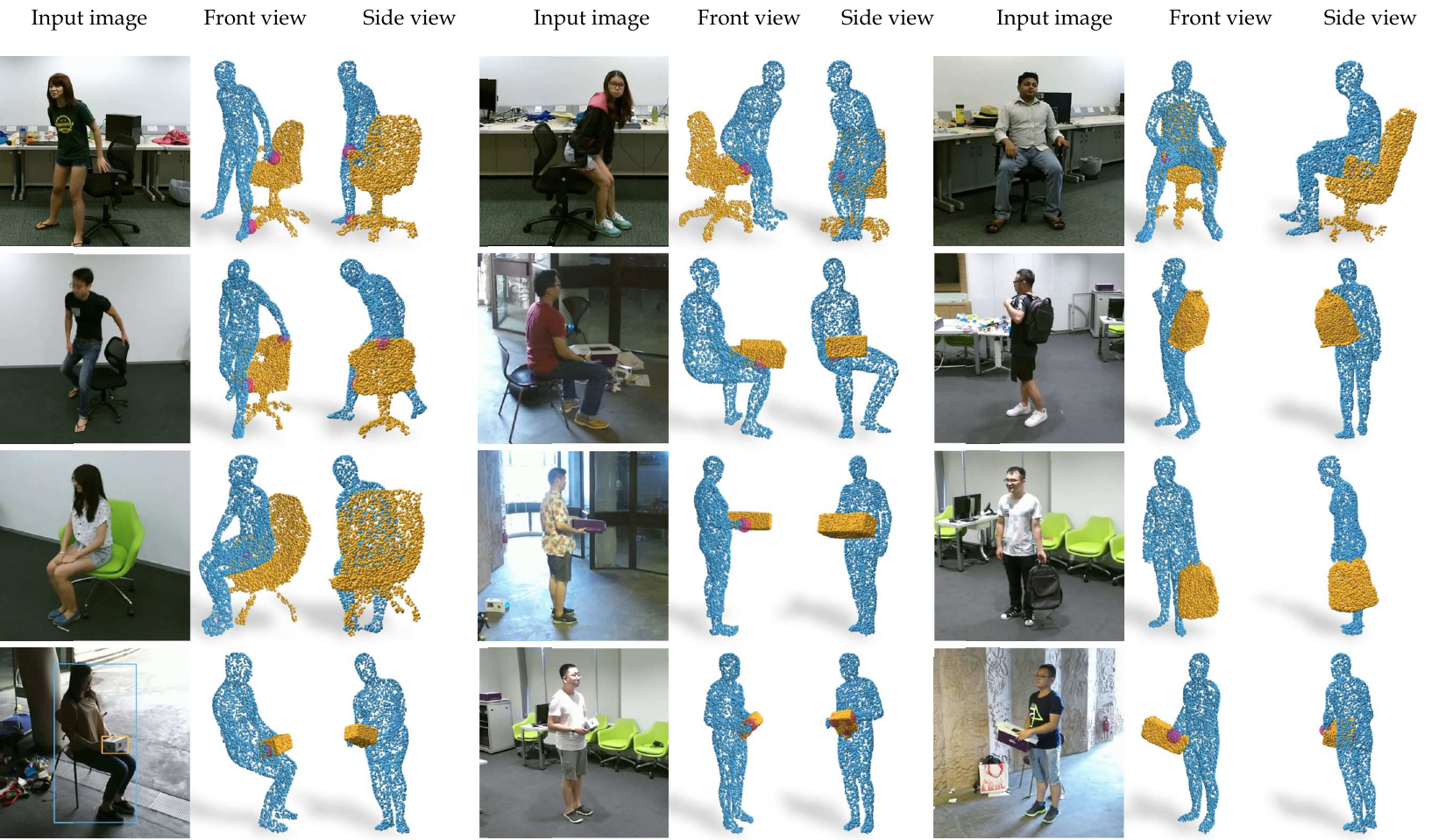}
    \caption{Generalization results on NTU-RGBD~\cite{Liu_2019_NTURGBD120} dataset. Our method can reconstruct different objects faithfully under various camera viewpoints and lighting conditions, without relying on any template shapes. }
    \label{fig:supp-ntu}
\end{figure*}

\begin{figure*}
    \centering
    \includegraphics[width=\linewidth]{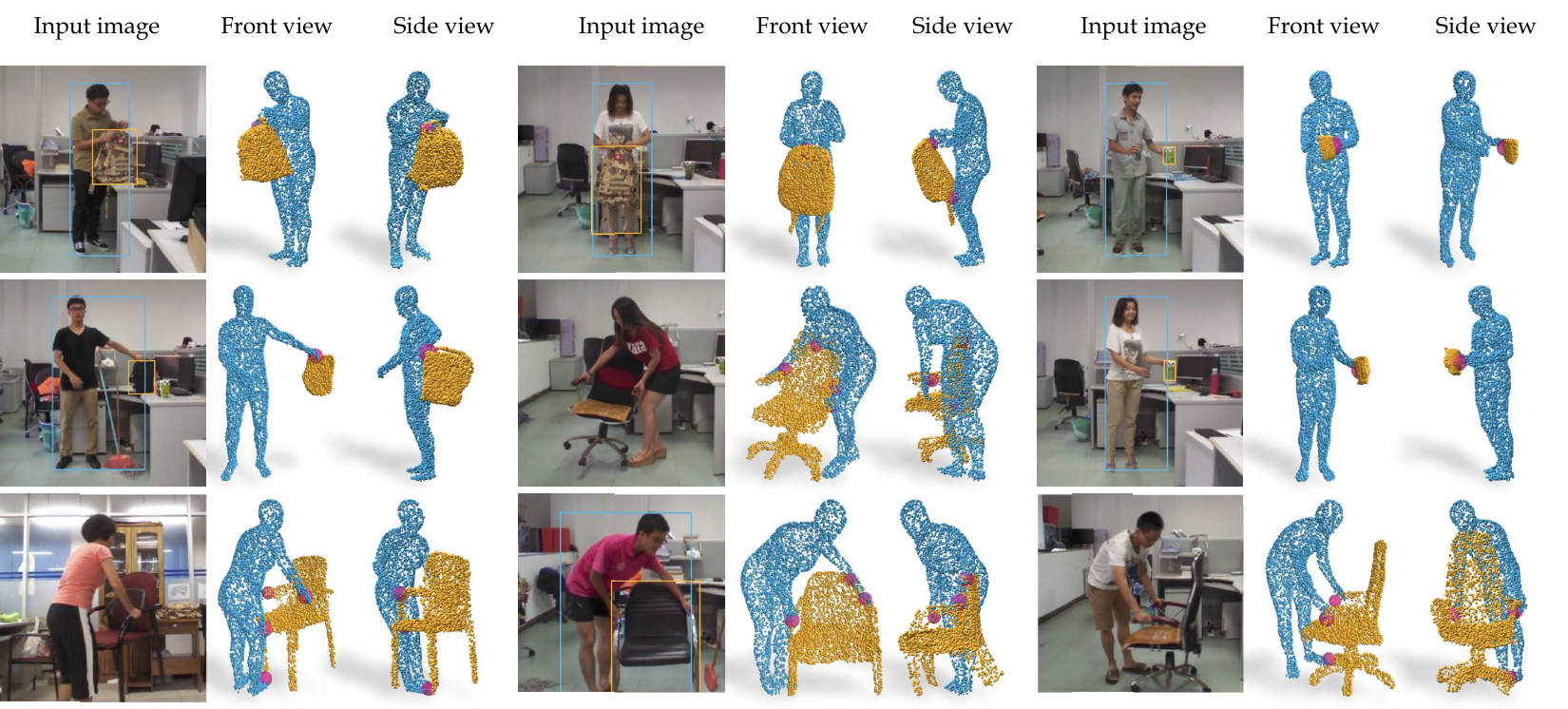}
    \caption{Generalization results on SYSU action~\cite{hu2017jointly_sysu} dataset. Our method can reconstruct different real-life human and objects during challenging interactions and occlusions. }
    \label{fig:supp-sysu}
\end{figure*}

\begin{figure*}
    \centering
    \includegraphics[width=\linewidth]{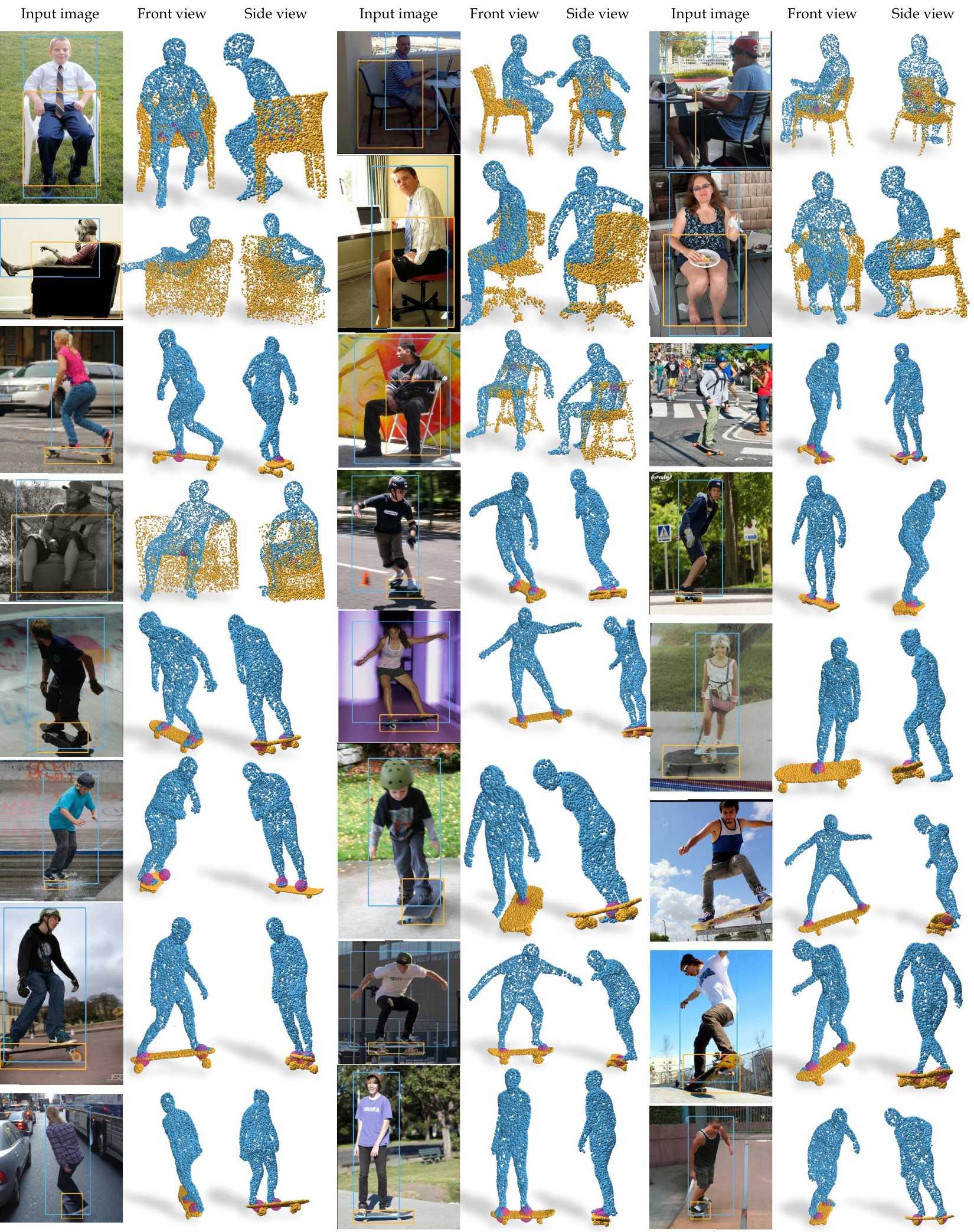}
    \caption{Generalization results to COCO~\cite{coco-dataset} dataset. Our method can reconstruct high-quality human and object from in the wild images which has very diverse shape variations, without using any template shapes.}
    \label{fig:supp-coco01}
\end{figure*}

\begin{figure*}
    \centering
    \includegraphics[width=\linewidth]{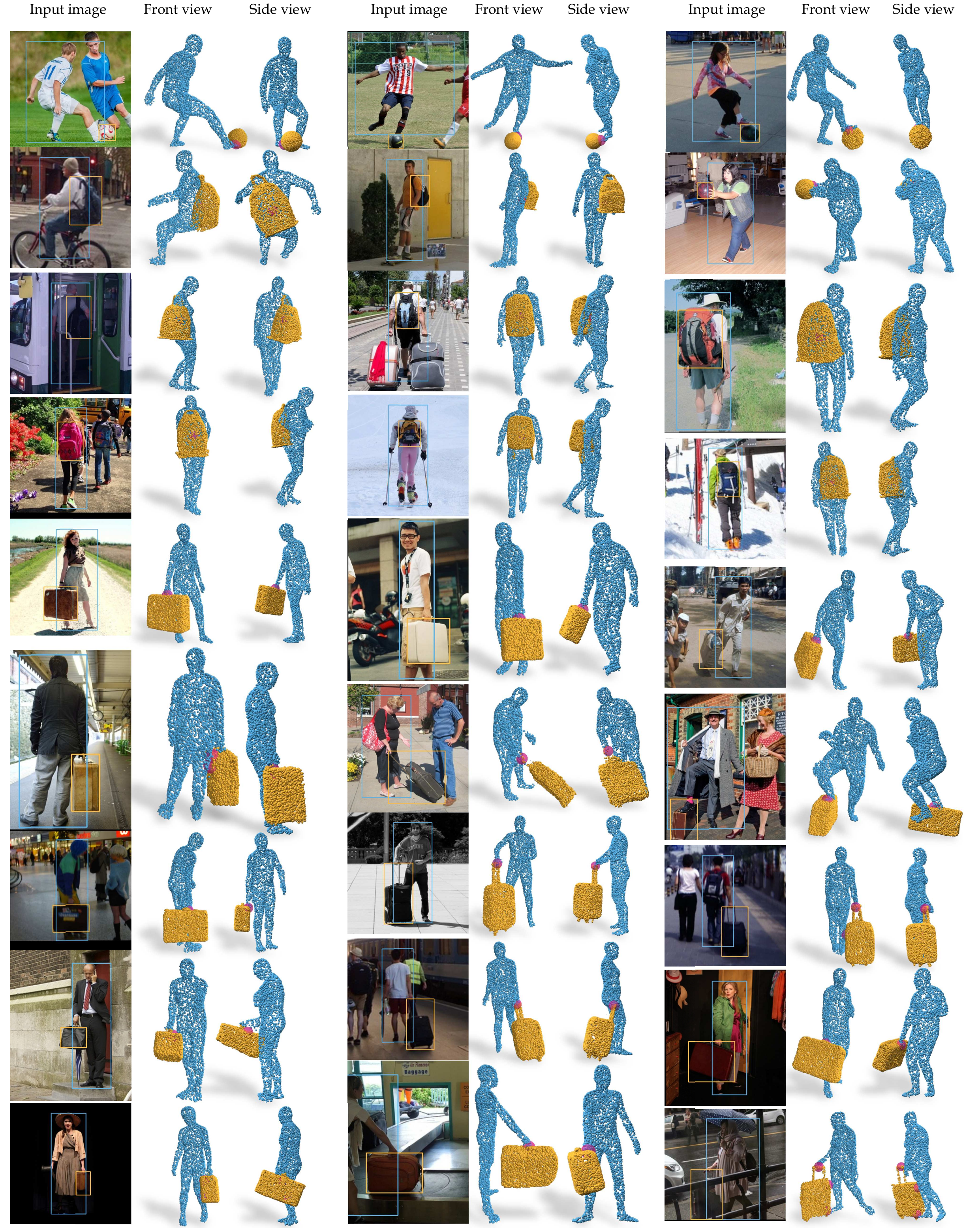}
    \caption{Generalization results to COCO~\cite{coco-dataset} dataset. Our method reconstructs diverse object shapes in the wild.}
    \label{fig:supp-coco02}
\end{figure*}

\begin{figure*}
    \centering
    \includegraphics[width=\linewidth]{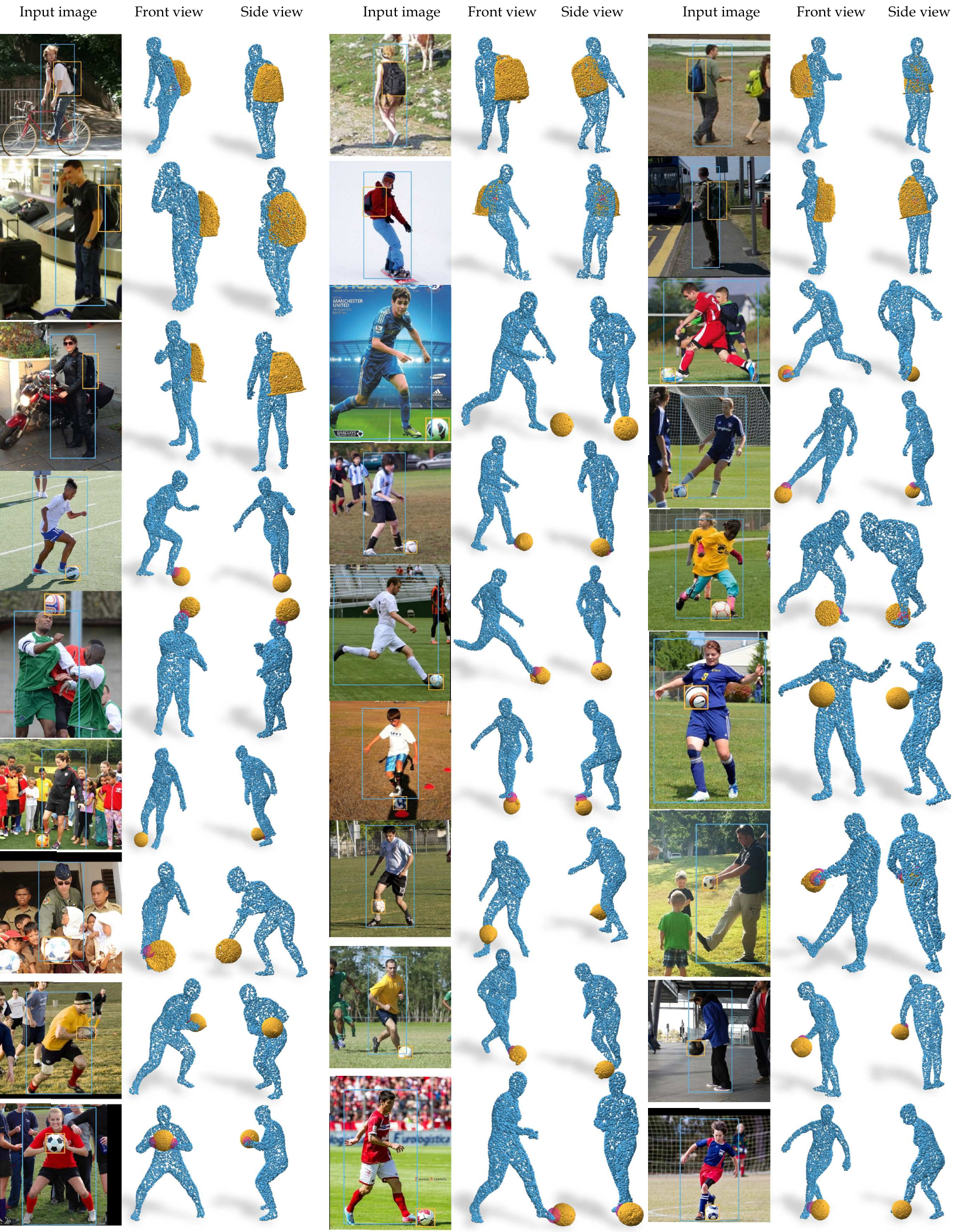}
    \caption{Generalization results to COCO~\cite{coco-dataset} dataset. Our method can reconstruct challenging human and object pose as well as shapes without using any template shapes.}
    \label{fig:supp-coco03}
\end{figure*}

Our method trained \emph{only} on our synthetic \dataName{} dataset generalizes well to other datasets. We show results on NTU-RGBD~\cite{Liu_2019_NTURGBD120}, SYSU~\cite{hu2017jointly_sysu} and challenging in the wild COCO~\cite{coco-dataset} images in figure \cref{fig:supp-ntu}, \cref{fig:supp-sysu} and \cref{fig:supp-coco01}, \cref{fig:supp-coco02}, \cref{fig:supp-coco03} respectively. Note that our method is trained only on our synthetic \dataName{} dataset and not fine-tuned on any images from these datasets. It can be seen that our method generalizes to different datasets with diverse object shapes, without requiring any template meshes. 

For quantitative evaluation, we focus on 15 object categories that are seen from our synthetic data (\cref{tab:category-fscore}). We test our method on three additional categories from BEHAVE and InterCap that are unseen and have GT data. The F-scores (human/object/combined) are: 0.465/0.453/0.479 (tennis racket), 0.333/0.332/0.361 (yoga mat), 0.375/0.305/0.360 (umbrella), 0.353/0.443/0.420 (all seen categories). We also show example reconstructions in \cref{fig:ood-generalization}. Our method can reconstruct \emph{unseen} categories.


\section{Limitations and Future Work}
We present a scalable solution to synthesize large amount of interaction dataset which allows training methods with strong generalization ability. We also propose a model for obtaining high quality human, object shapes and also interaction semantics, without any template shapes. We demonstrate the generalization ability of our method on diverse datasets. Our template-free reconstruction method is a promising first step towards real in-the-wild reconstruction. 

Nevertheless, there are still some limitations to the current approach. First, our \dataName{} data generation method always starts with a seed interaction pose sampled from an existing interaction dataset. This limits the diversity in terms of interaction poses. Future works can explore generative models such as Object-Popup~\cite{petrov2020objectpopup} to further diversify the interaction pose. It is also highly desirable to combine the large human pose variations from AMASS~\cite{AMASS:ICCV:2019}, which can further improve the robustness of reconstruction methods to challenging poses. 

Secondly, our method struggles to predict accurate human shapes when large chunk of the human body is occluded, see \cref{fig:supp-failure}. This is because our method is purely template-free and only use the network to learn the human and object shape priors. Future works can try to further explore human shape or pose constraints to regularize network training and predictions. In addition, our hierarchical diffusion model are designed for human object interaction, which is applicable for general bilateral interaction cases like human-human, hand-hand, and hand-object interactions. However, it cannot handle multi-person or multi-object interactions. We leave these for future works.

\end{document}